\documentclass{bmvc2k}
\usepackage{graphicx}
\usepackage{float}
\usepackage{todonotes}
\usepackage{subfigure}

\title{Wildest Faces: Face Detection and Recognition in Violent Settings}

\addauthor{Mehmet Kerim Yucel$^{*}$}{mkerimyucel@hacettepe.edu.tr}{1}
\addauthor{Yunus Can Bilge$^{ * }$}{yunuscan.bilge@hacettepe.edu.tr}{1}
\addauthor{Oguzhan Oguz}{oguzhan.oguz@hacettepe.edu.tr}{1}
\addauthor{Nazli Ikizler-Cinbis}{nazli@cs.hacettepe.edu.tr}{1}
\addauthor{Pinar Duygulu}{pinar@cs.hacettepe.edu.tr}{1}
\addauthor{Ramazan Gokberk Cinbis}{gcinbis@ceng.metu.edu.tr}{2}

\addinstitution{
 Department of Computer Science\\
 Hacettepe University\\
 Ankara, Turkey \\
}
\addinstitution{
 Department of Computer Engineering\\
 Middle East Technical University\\
 Ankara, Turkey \vspace{3mm}  \\
 * indicates equal contribution. 
}

\runninghead{Yucel et al.}{Wildest Faces}

\begin{document}

\maketitle

\begin{abstract}

With the introduction of large-scale datasets and deep learning models capable of learning complex representations, impressive advances have emerged in face detection and recognition tasks.
Despite such advances, existing datasets do not capture the difficulty of face recognition in the \textit{wildest} scenarios, such as hostile disputes or fights. Furthermore, existing datasets do not represent completely unconstrained cases of low resolution, high blur and large pose/occlusion variances. To this end, we introduce the \textit{Wildest Faces} dataset, which focuses on such adverse effects through violent scenes. The dataset consists of an extensive set of violent
scenes of celebrities from movies. 
Our experimental results demonstrate that state-of-the-art
techniques are not well-suited for violent scenes, and therefore, \textit{Wildest Faces} is likely to stir further interest in face detection and recognition research.

\end{abstract}

\section{Introduction}
\label{sec:intro}

Detection and recognition of faces have a wide range of application areas, such as surveillance, consumer products and security systems. With the emergence of deep learning, impressive accuracies have been reported in face detection~\cite{li2015convolutional,farfade2015multi,najibi2017ssh, hu2017finding} compared to earlier results obtained by hand-crafted feature pipelines such as~\cite{viola2004robust, yang2014aggregate,li2013learning, mathias2014face}. Example approaches include cascade systems for multi-scale detection~\cite{li2015convolutional,yang2016wider,zhang2016joint}, facial-part scoring~\cite{yang2015facial,samangouei2018face}, proposal-stage anchor design~\cite{zhang2017s,zhu2018seeing}, ensemble systems~\cite{hu2017finding,yang2017face}, optimized single-stage detectors~\cite{najibi2017ssh,tang2018pyramidbox} and integrated attention mechanisms\cite{wang2017face} (see the survey \cite{zafeiriou2015survey}).

Likewise, there has been a plethora of studies on face recognition.  Compared to the pioneering works of \cite{turk1991face, ahonen2006face, xie2010fusing,edwards1998face, wright2009robust, wiskott1997face}, face recognition models that benefit from deep learning-based techniques and concentrate on better formulation of distance metric optimization raised the bar~\cite{schroff2015facenet,taigman2014deepface,parkhi2015deep, wen2016discriminative,sun2013hybrid,sun2014deep,sun2015deepid3}. In addition to face recognition in still images, video-based face recognition studies have also emerged (see  \cite{ding2016comprehensive} for a recent survey). Ranging from local feature-based methods~\cite{li2013probabilistic,parkhi2014compact,li2014eigen} to manifolds \cite{huang2015log} and metric learning~\cite{cheng2018duplex,huang2017cross,goswami2017face}, recent studies have focused on finding informative frames in image sets \cite{goswami2014mdlface} and finding efficient and fast ways of feature aggregation~\cite{chowdhury2016one,yang2017neural,rao2017learning, rao2017attention}.

Nevertheless, the real-life conditions still challenge the state-of-the-art algorithms due to variations in scale, background, pose, expression, lighting, occlusion, age, blur and image resolution. As shown in \cite{yang2016wider}, several leading algorithms produce severely degraded results in rather unconstrained conditions. Recently, there have been many attempts in building large scale datasets with variety of real-life conditions. FDDB \cite{jain2010fddb}, AFW \cite{zhu2012face}, PASCAL Faces \cite{yan2014face}, Labeled Faces in the Wild (LFW) \cite{huang2007labeled}, Celeb Faces \cite{sun2013hybrid}, Youtube Faces  (YTF) \cite{wolf2011face}, IJB-A \cite{klare2015pushing}, MS-Celeb-1M \cite{guo2016ms}, VGG-Face \cite{parkhi2015deep},  VGG2-Face \cite{cao2017vggface2}, MegaFace \cite{kemelmacher2016megaface} and WIDER Face \cite{yang2016wider} datasets have been made publicly available for research purposes. Datasets with extreme scales, such as \cite{schroff2015facenet} and \cite{taigman2014deepface} have also been used but have not been disclosed to the public.
However, these datasets can still be considered as "controlled" in several regards, such as resolution, the presence of motion blur and the very quality of the image. Moreover, these datasets mostly omit noisy samples and are not representative of extreme expressions, such as anger and fear in violent scenes.

\begin{figure}%
\centering
\subfigure{%
\label{}%
\includegraphics[width=0.7\linewidth]{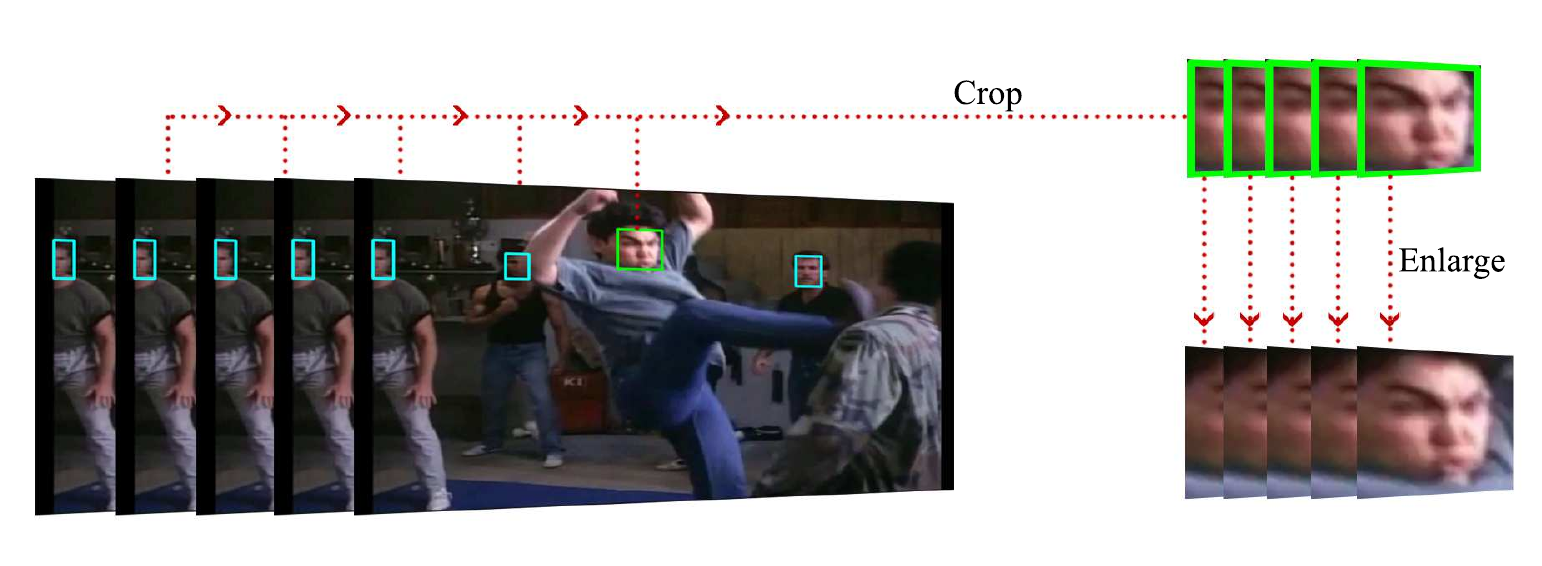}}%
\vspace{-4mm}
\\
\subfigure{%
\label{}%
\includegraphics[width=0.09\linewidth]{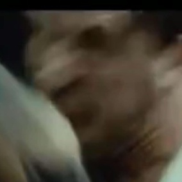}}%
\subfigure{%
\label{}%
\includegraphics[width=0.09\linewidth]{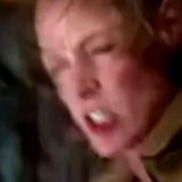}}%
\subfigure{%
\label{}%
\includegraphics[width=0.09\linewidth]{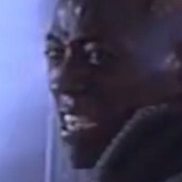}}%
\subfigure{%
\label{}%
\includegraphics[width=0.09\linewidth]{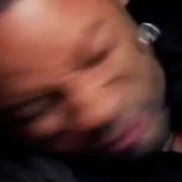}}%
\subfigure{%
\includegraphics[width=0.09\linewidth]{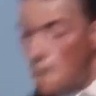}}%
\subfigure{%
\label{}%
\includegraphics[width=0.09\linewidth]{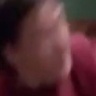}}
\subfigure{%
\label{}%
\includegraphics[width=0.09\linewidth]{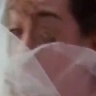}}%
\subfigure{%
\label{}%
\includegraphics[width=0.09\linewidth]{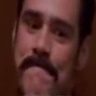}}%
\subfigure{%
\label{}%
\includegraphics[width=0.09\linewidth]{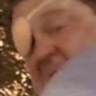}}%
\subfigure{%
\includegraphics[width=0.09\linewidth]{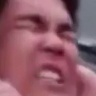}}%
\label{fig:exampleImages}%
\vspace{-3mm}
\caption{ Our dataset creation pipeline is shown in the first row. Faces with green bounding boxes indicate the celebrities that are used for  recognition. Second row shows sample recognition images from \textit{Wildest Faces} dataset which include variety of real-life conditions. Note the amount of pose variations, blur and low image quality. Moreover, \textit{Wildest Faces} offers a considerable age variance, extreme facial expressions as well as severe occlusion.}
\vspace{-6mm}
\end{figure}

In this paper, we present a new benchmark dataset, namely \textit{Wildest Faces}, where we put the emphasis on violent scenes with virtually unconstrained scenarios. In addition to previously studied adverse conditions, \textit{Wildest Faces} dataset contains images from a large spectrum of image quality, resolution and motion blur (see Fig. \ref{fig:exampleImages}).
The dataset consists of videos of celebrities in which they are practically fighting. There are $ \sim 68$K  images (a.k.a frames) and 2186 shots of 64 celebrities, and all of the video frames are  manually annotated to foster research both for detection and recognition of \textit{``faces in the wildest''}. It is especially important from the surveillance perspective to identify the people who are involved in crime scenes and we believe that the availability of such a dataset of violent faces would stir further research towards this direction as well.

We provide a detailed discussion of the statistics and the evaluation of state-of-the-art methods on the proposed dataset. We exploit the dataset both in the context of face detection, image-based and video-based face recognition. For video face recognition, we also introduce an attention-based temporal pooling technique to aggregate videos in a simple and effective way. Our experimental results demonstrate that such a technique can be preferable amongst others, whilst there is still a large room for improvement in this challenging dataset that is likely to facilitate further research. 
\vspace{-4mm}

\section{Discussion on available datasets} \label{related_work}

\textbf{Face Detection Datasets:}
AFW \cite{zhu2012face} contains background clutter with different face variations and associated annotations include bounding box, facial landmarks and pose angle labels. FDDB~\cite{jain2010fddb} is built using Yahoo!, where images with both eyes in clear sight are neglected, which leads to a rather constrained distribution in terms of pose and occlusion. IJB-A \cite{klare2015pushing} is one of the few datasets that contains annotations for both recognition and detection tasks. MALF \cite{yang2015fine} incorporates rich annotations in the sense that they contain pose, gender and occlusion information as well as expression information with a certain level of granularity. PASCAL Faces \cite{yan2014face} contains images selected from PASCAL VOC \cite{everingham2010pascal}. In AFLW \cite{koestinger11a} annotations come with rich facial landmark information available. WIDER Face \cite{yang2016wider} is one of the largest datasets released for face detection. Collected using categories chosen from LSCOM \cite{naphade2006large}, each annotation is categorized due to its scale, occlusion, pose, overall difficulty and events, which facilitates in-depth analysis.  Detailed information on these datasets can be found in Table \ref{detection_datasets}.

\begin{table}[]
\tiny
\resizebox{\textwidth}{!}{%
		\begin{tabular}{|l|l|l|l|l|l|}
			\hline
			\textbf{Dataset} & \textbf{\# Images} & \textbf{\# Faces} & \textbf{Source} & \textbf{Type} &  \textbf{Public} \\
			\hline\hline

			AFW \cite{zhu2012face} & 205  & 473 & Flickr & Images &  Yes  \\ \hline

			FDDB \cite{jain2010fddb} &  2,845  &  5,171  & Yahoo! News & Images & Yes \\ \hline

			IJB - A \cite{klare2015pushing} & 24,327  & 49,579 & Internet & Images / Videos & Yes  \\ \hline

			MALF \cite{yang2015fine} & 5,250  & 11,931  & Flickr, Baidu Inc. & Images  & Yes  \\ \hline

			AFLW \cite{koestinger11a} & 21,997  & 25,993  & Flickr & Images  & Yes  \\ \hline

			PASCAL Faces \cite{yan2014face} & 851  & 1,335  & PASCAL VOC & Images  & Yes  \\ \hline

			WIDER Face  \cite{yang2016wider}& 32,203 & 393,703   & Google, Bing & Images  & Yes  \\ \hline
			\textbf{Wildest Faces} & 67,889 & 109,771 & YouTube & Videos & Yes \\ \hline

		\end{tabular}%
}
\vspace{-4mm}
	\caption{Face detection datasets.}
\label{detection_datasets}
\vspace{-4mm}
\end{table}

\vspace{2mm}
\noindent \textbf{Face Recognition Datasets:}
Labeled Faces in the Wild (LFW) \cite{huang2007labeled} is one of the widely used datasets in the recognition literature. Viola-Jones detector \cite{viola2001rapid} is used to detect faces during the dataset collection phase, and then manual correction on annotations is performed. PubFig \cite{kumar2009attribute} is created as a complement to the LFW. The faces in this set are the images of the public celebrities and are collected using Google and Flickr. Celebrity Faces \cite{sun2013hybrid} is constructed using public figures. In one of the turning points of face recognition, large-scale VGG face dataset \cite{parkhi2015deep} is released with the help of automated face detection and a stunning number of 200 human annotators. During its collection phase, care is taken to avoid having the same individuals with LFW and YTF datasets. Recently, this dataset is further expanded in  \cite{cao2017vggface2} as VGG Face-2, which is fairly larger than its predecessor.  FaceScrub \cite{ng2014data} is another dataset comprised of individuals who are primarily celebrities. CASIA-WebFace \cite{yi2014learning} is another popular dataset, though authors note that they can't be sure that all images are annotated correctly. MS-Celeb-1M \cite{guo2016ms} contains approximately 10 million images of 100,000 individuals where 1,500 of them are celebrities. In one of the latest benchmarks released publicly, MegaFace \cite{kemelmacher2016megaface} contains a large set of pictures from Flickr with a size of 50 pixels in both dimensions, where faces are detected using Headhunter \cite{mathias2014face}. Authors of \cite{kemelmacher2016megaface} also presented an improved version of MegaFace, dubbed MF2 \cite{nech2017level}, that builds on its predecessor. Additionally, tech giants have utilized their proprietary datasets in Facebook's DeepFace \cite{taigman2014deepface}, Google's FaceNet \cite{schroff2015facenet} and NTechLab's \footnote{\url{https://ntechlab.com }}.

For video face recognition, YouTube Faces \cite{wolf2011face} uses \cite{viola2001rapid} to automatically detect faces. Each face in the data is centered, expanded with 2.2 magnification factor and the size of the annotation is fixed with 100 pixels in both dimensions. Other two prominent video face recognition datasets are COX \cite{huang2015benchmark} and PasC \cite{beveridge2013challenge}. Despite their relatively large size, PasC \cite{beveridge2013challenge} suffers from video location constraints and COX \cite{huang2015benchmark} suffers from demographics as well as video location constraints. Detailed information on these datasets can be found in Table~\ref{recognition_sets}.

\vspace{2mm}
\noindent\textbf{Limitations of the available datasets:}
Except WIDER, the available datasets generally focus on high resolution and high quality images. Moreover, several of these datasets filter low quality, occluded and blurred images, thus do not represent what is out there in the real world. Although there are video recognition datasets  which inherently consist of motion blurred or comparably low quality images (e.g. \cite{wolf2011face}), majority of the datasets are likely to suffer from automatically performed face detector bias. In addition, to the best of our knowledge, none of these datasets primarily focus on violent scenes where unconstrained scenarios might actually introduce unconstrained effects.

\vspace{-4mm}

\begin{table}[t]
\tiny
	\begin{center}
\resizebox{\textwidth}{!}{%
		\begin{tabular}{|l|l|l|l|l|}
			\hline
			\textbf{Dataset} &  \textbf{\# Images (or videos)} & \textbf{\# Individuals} & \textbf{Source} & \textbf{Type} \\
			\hline\hline

			\textbf{Wildest Faces} & 2,186 (64,242 frames)  & 64 & YouTube  & Videos  \\ \hline

			COX \cite{huang2015benchmark} & 3,000 & 1000 & Custom & Videos \\ \hline
			PasC \cite{beveridge2013challenge} & 2,802 + 9376 frames &293 &  Custom & Videos \\ \hline 	
			YTF \cite{wolf2011face} & 3,425  & 1,595  & Youtube &  Videos \\ \hline \hline		
			LFW \cite{huang2007labeled} & 5,749  & 13,233 & Yahoo! News & Images  \\ \hline

			PubFig \cite{kumar2009attribute} &  60,000  &  200  & Google, Flickr & Images \\ \hline

			CelebA \cite{yang2015facial} & 202,599  & 10,177  & Google, Bing & Images  \\ \hline

			CelebFaces \cite{sun2013hybrid} & 87,628  & 5,436  & Flickr, Baidu Inc. & Images \\ \hline

			VGG Face \cite{parkhi2015deep} & 2.6M  & 2,622  & Google, Bing & Images \\ \hline

			FaceScrub \cite{ng2014data} & 106,863  & 530  & Internet & Images  \\ \hline

			CASIA-WebFace \cite{yi2014learning} & 494,414 & 10,000  & IMDB & Images \\ \hline

			MegaFace \cite{kemelmacher2016megaface} & 1M & 690,572   & Flickr & Images  \\ \hline

			VGG-2 \cite{cao2017vggface2} & 3.2M & 9,131   &  Google  & Images  \\ \hline

			MF2 \cite{nech2017level} & 4.7M & 672,000   & Flickr & Images \\ \hline

			MS-Celeb-1M  \cite{guo2016ms} & 10M & 100,000  & Internet,Bing & Images \\ \hline

			DeepFace \cite{taigman2014deepface}\dag	 & 4M & 4,000  &  Internal & Images \\ \hline

			FaceNet  \cite{schroff2015facenet} \dag	& 500 M & 8 M  &  Internal & Images  \\  \hline

			NTechLab \dag{} & 18.4 M & 200.000  &  Internal & Images  \\
			\hline
		\end{tabular}
}
	\end{center}
	\vspace{-2mm}
	\caption{Face recognition datasets. \dag{} indicates private dataset. Among the available video face recognition datasets, \textit{Wildest Faces} have the highest video count per individual.}
\label{recognition_sets}
\vspace{-6mm}
\end{table}

\section{Wildest Faces Dataset} \label{dataset_info}

\begin{figure}%
\centering
\subfigure{%
\includegraphics[width=0.1\linewidth]{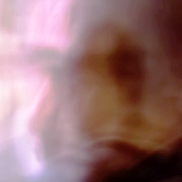}}%
\subfigure{%
\includegraphics[width=0.1\linewidth]{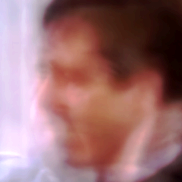}}%
\subfigure{%
\includegraphics[width=0.1\linewidth]{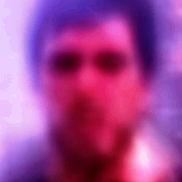}}%
\subfigure{%
\includegraphics[width=0.1\linewidth]{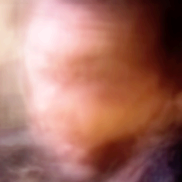}}%
\subfigure{%
\includegraphics[width=0.1\linewidth]{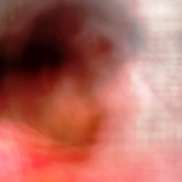}}%
\subfigure{%
\includegraphics[width=0.1\linewidth]{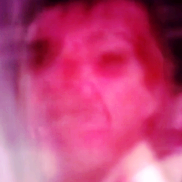}}%
\subfigure{%
\includegraphics[width=0.1\linewidth]{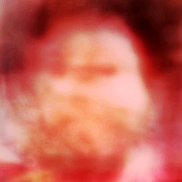}}%
\subfigure{%
\includegraphics[width=0.1\linewidth]{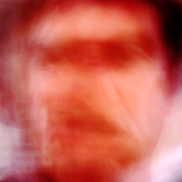}} \\
\subfigure{%
\includegraphics[width=0.1\linewidth]{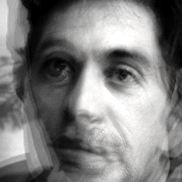}}%
\subfigure{%
\includegraphics[width=0.1\linewidth]{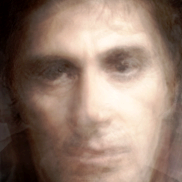}}%
\subfigure{%
\includegraphics[width=0.1\linewidth]{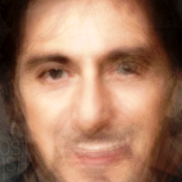}}%
\subfigure{%
\includegraphics[width=0.1\linewidth]{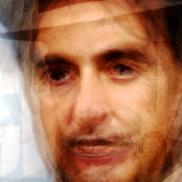}}%
\subfigure{%
\includegraphics[width=0.1\linewidth]{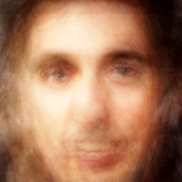}}%
\subfigure{%
\includegraphics[width=0.1\linewidth]{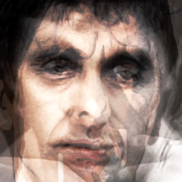}}%
\subfigure{%
\includegraphics[width=0.1\linewidth]{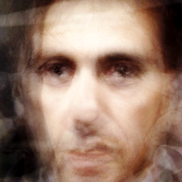}}%
\subfigure{%
\includegraphics[width=0.1\linewidth]{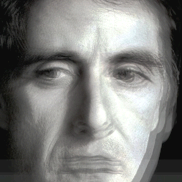}} \\
\subfigure{%
\includegraphics[width=0.1\linewidth]{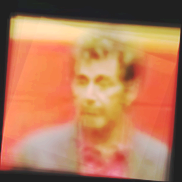}}%
\subfigure{%
\includegraphics[width=0.1\linewidth]{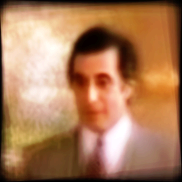}}%
\subfigure{%
\includegraphics[width=0.1\linewidth]{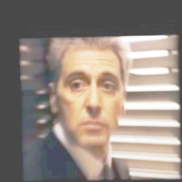}}%
\caption{K-Means cluster centers  for Al Pacino images in \textit{Wildest Faces}, FaceScrub \cite{ng2014data} and YouTube Faces \cite{wolf2011face} are shown in first, second and third row, respectively. k=8 for \textit{Wildest Faces} and FaceScrub , k=3 for YouTube Faces as higher k values produce repetitive images. Average faces from \textit{Wildest Faces} are the least recognizable, indicating a large degree of variance in adverse effects. Images are histogram equalized for convenience.}
\label{k_means_images}
\vspace{-4mm}
\end{figure}

Human faces are in their wildest form during violence or fight with their expressions uncontrolled. Besides, the fast movements during violence naturally results in challenges for pose, occlusion, and blur. Based on these observations, we constructed \textit{Wildest Faces} dataset from YouTube videos by focusing on violent scenes of celebrities in movies.

\vspace{-2mm}
\subsection{Data Collection and Annotation}
We first identified the celebrities who are known to be acting in movies with violence. We then picked their videos from YouTube in a variety of scene settings; car chase, indoor fist fights, gun fights, heated arguments and science fiction/fantasy battles. This abundance in scene settings provide an inherent variety of possible occluding objects, poses, background clutter and blur (see Fig.\ref{fig:exampleImages}).  Majority of the frames of each video have celebrity face in them, though in some frames celebrities may not be present. Videos, with an average 25 FPS are then divided into shots with a maximum duration of 10 seconds.

In total, we choose 64 celebrities and collect 2,186 shots from 410 videos, which results in 67,889 frames with 109,771 manually annotated bounding boxes. In order to test the generalization ability thoroughly, we split the dataset based on videos and do not include any shots from a training video in the other splits.  The splits for training, validation and test sets yield the ratios 56\%-23\%-21\% video-wise and 61\%-20\%-19\% frame-wise. Video-based splitting also assesses age difference; e.g. training set includes Sean Connery in his early acting days whereas test set solely includes him in late stages of his career.

Ground truth locations of faces have been annotated by 12 annotators using VOTT \footnote{\url{https://github.com/Microsoft/VoTT}}. We also label our celebrities with \textit{target} tag for recognition and label the rest of the faces as \textit{non-target}. We do not omit any adverse effect; we label extremely tiny, occluded, frontal/profile and blurred faces. When creating the recognition set, we simply crop the target label from each frame in the dataset and expand the area with a factor 0.15 to make sure we do not miss any facial parts. An example illustration can be seen in Fig.\ref{fig:exampleImages}. As we do not have celebrity faces in each collected frame, our recognition set consists of 64,242 frames in total.

\vspace{-2mm}
\subsection{Statistics}

\begin{figure}%
\centering
\subfigure[Detection scales. ]{%
\label{scale_det}%
\includegraphics[width=0.24\linewidth]{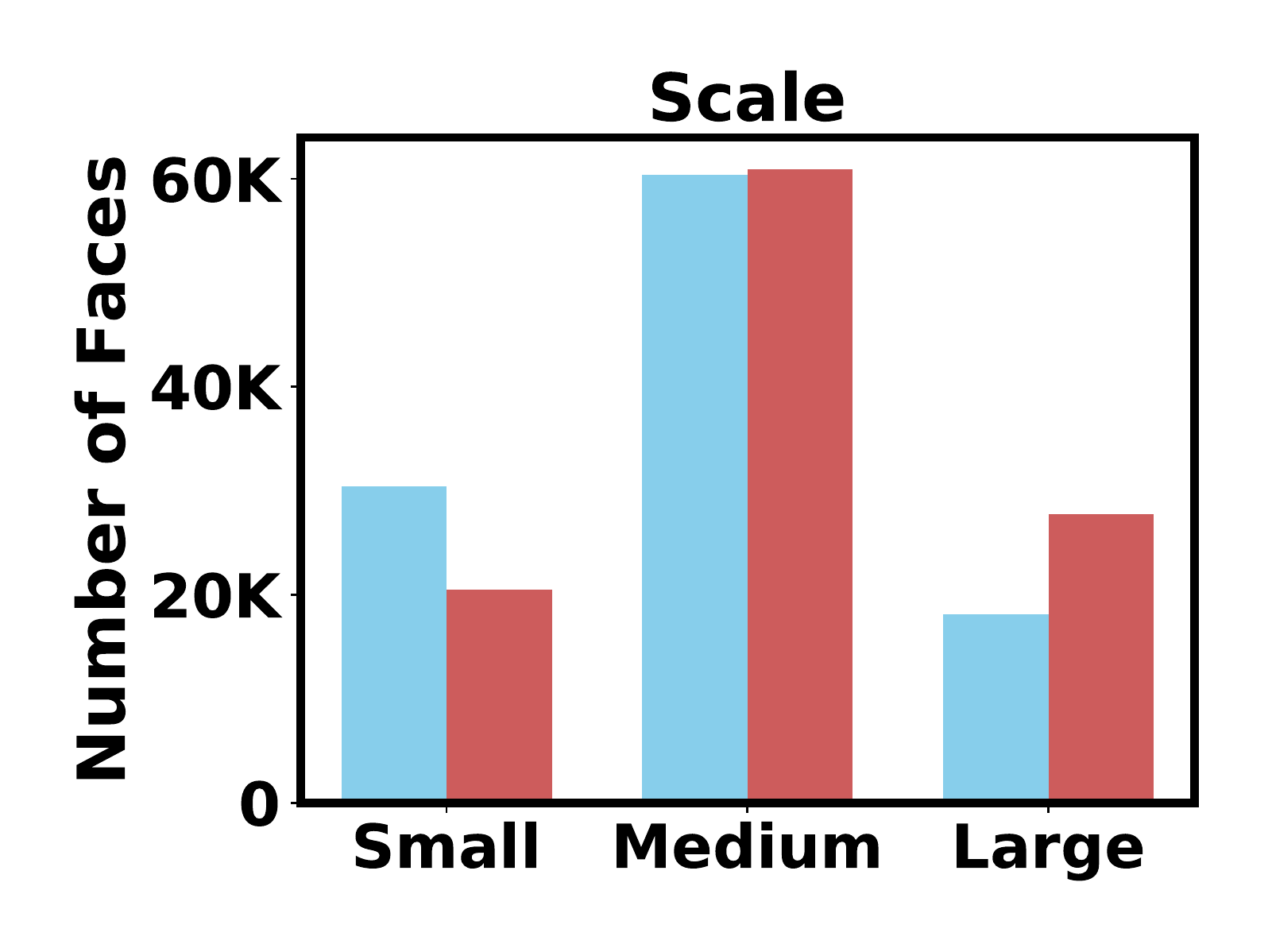}}%
\subfigure[Detection blur.]{%
\label{blur_det}%
\includegraphics[width=0.24\linewidth]{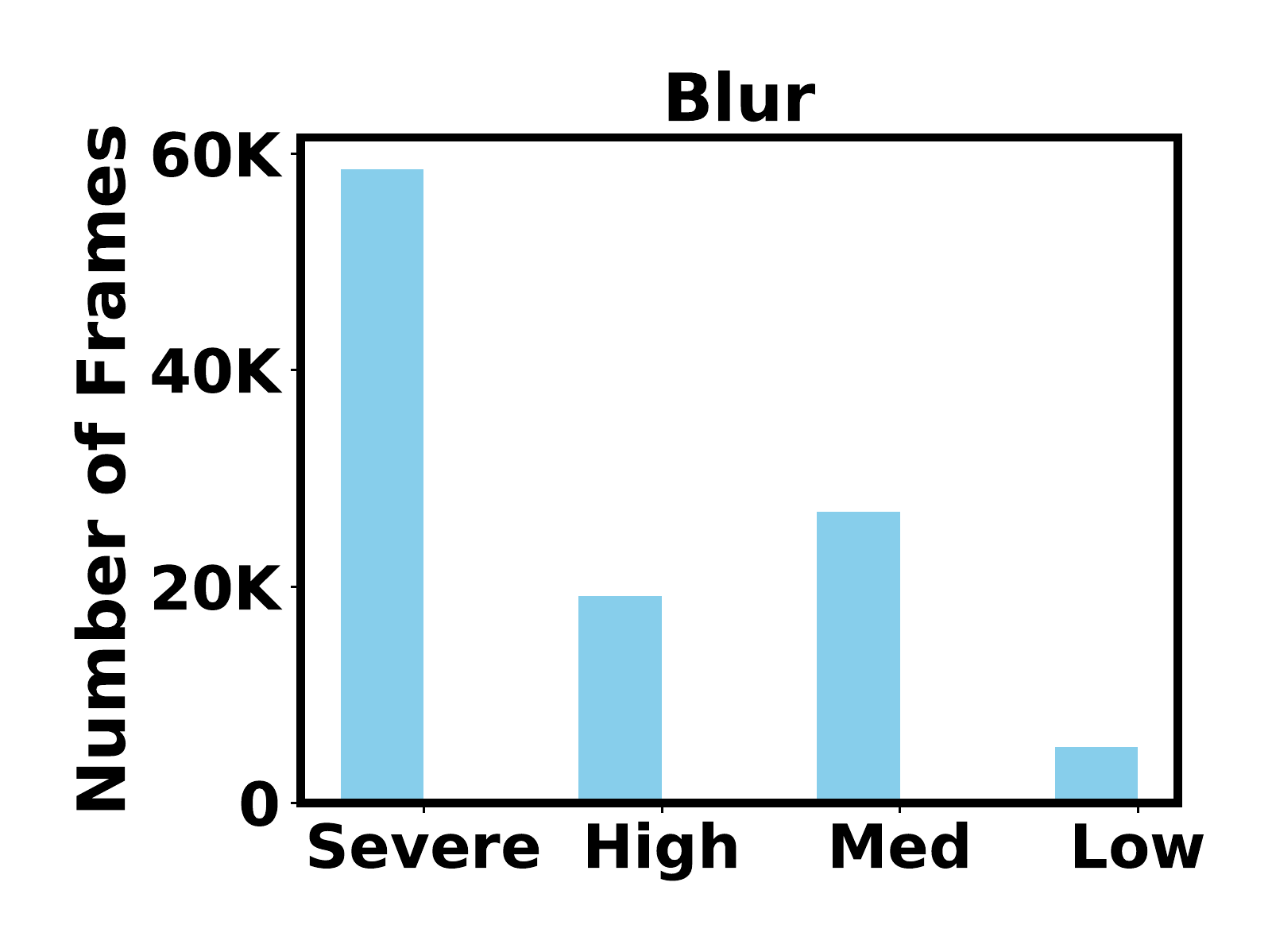}}%
\subfigure[Recognition blur.]{%
\label{blur_rec}%
\includegraphics[width=0.24\linewidth]{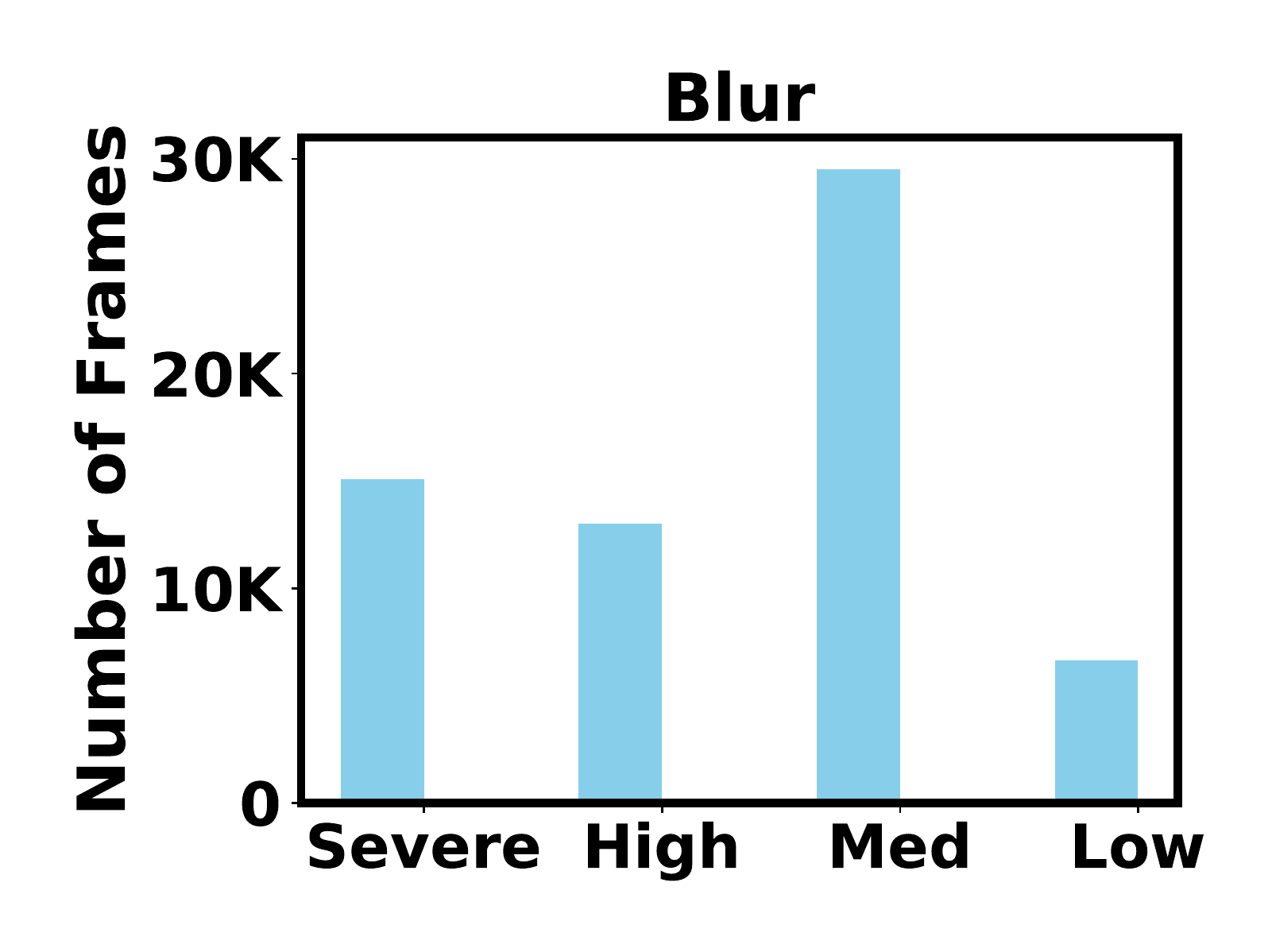}}%
\subfigure[Recognition age.]{%
\label{age_variance}%
\includegraphics[width=0.23\linewidth]{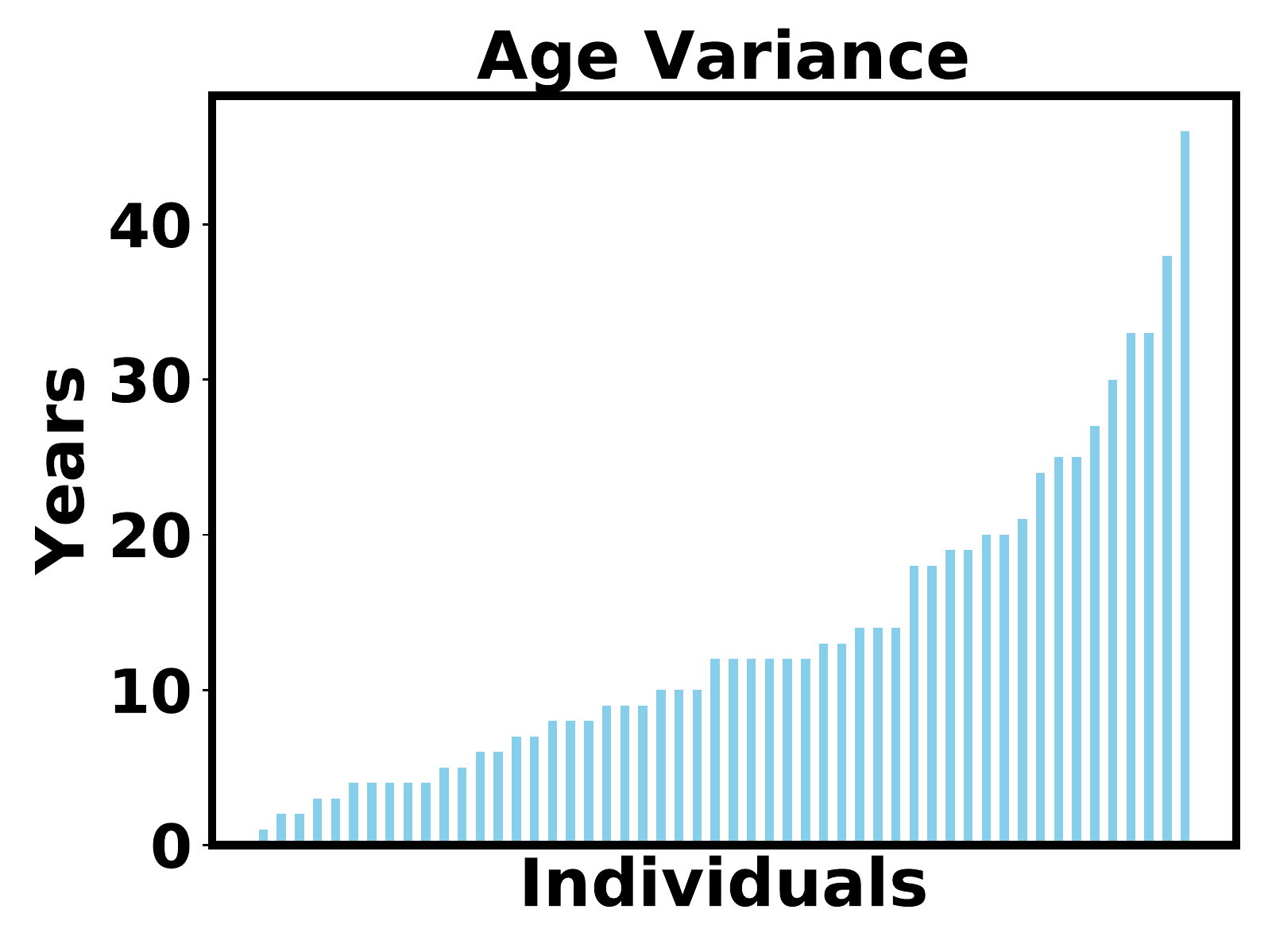}}%
\caption{\textit{Wildest Faces} Statistics. In  (a),  blue  and  red  correspond  to  width  and  height, respectively. Detection set offers a severely blurred data, whereas recognition has a more equal distribution. For detection scales, we see an equal emphasis on small and large faces.  }
\vspace{-6mm}
\end{figure}

Wildest Faces dataset has a diverse distribution of faces. In Figure~\ref{k_means_images}, k-means cluster centers of Al Pacino's images (dataset-wide) are shown for FaceScrub \cite{ng2014data}, YouTubeFaces \cite{wolf2011face} and Wildest Faces. It is clear that our dataset has a wide spectrum of adverse effects as its cluster centers are far from being recognizable as Al Pacino. Wildest Faces offers a good scale variance for detection, as well as high amount of blur. Recognition set offers a good distribution of several blur levels as well as a noticable average age variance. Occluded shots roughly makes up the half of the available data, which offers a challenge as well. Moreover, pose variance is sufficiently large in each shot, which would promote pose-invariance in video face recognition. In the following, we present the analysis of these effects.

\noindent\textbf{Scale.} We classify our faces into categories of \textit{small}, \textit{medium} and \textit{large} with respect to the heights of faces: below 100 pixels as \textit{small}, in between 100 to 300 pixels as \textit{medium}, and larger than 300 pixels as \textit{large}. Scale statistics for detection set is shown in Figure \ref{scale_det}. For recognition set, the balance shifts slightly to \textit{medium} from \textit{small}.

\noindent\textbf{Blur.} We follow a multi-stage procedure to quantify blur that is present in images. Inspired from \cite{pech2000diatom}, we perform contrast normalization and then convert our images to grayscale. Grayscale images are then convolved with a 3x3 Laplacian Kernel, and variance of the result is used to produce a blurness value, which is then used to empirically find a threshold to divide the images into blur categories.. We then manually edit any wrong blur labels. Blur statistics are shown in Figure~\ref{blur_det} and Figure~\ref{blur_rec}.

\noindent\textbf{Age.} For each individual we also measure the distribution of age variances, which is the differences between the dates of their earliest and latest movies in our dataset. We see drastic age variations in certain individuals, up to 40 years (see Figure~\ref{age_variance}). On average, 13 years of age variation per individual is observed.

\noindent\textbf{Occlusion.} We provide occlusion information on shot-level for recognition set; we label shots as \textit{no occlusion}, \textit{mixed} or \textit{significant}. Shots labelled \textit{mixed} have occlusion in several frames of the shot, but not more than half of the face is occluded. \textit{Significant} labels indicate there are several frames with heavy occlusion, where at least half of the face is occluded.  We randomly select 250 shots in our dataset and analyze them; this leads to a ratio of 20\%, 28\% and 52\% for \textit{significant}, \textit{mixed} and \textit{no occlusion} tags, respectively.

\begin{figure}%
\centering
\subfigure[Al Pacino]{%
\label{al_lblp}%
\includegraphics[width=0.1\linewidth]{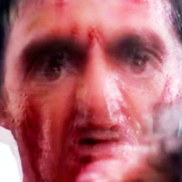}%
\label{al_lbhp}%
\includegraphics[width=0.1\linewidth]{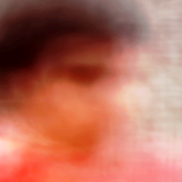}}%
\hspace{2mm}%
\subfigure[Dwayne Johnson]{%
\label{dj_hphb}%
\includegraphics[width=0.1\linewidth]{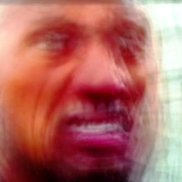}%
\label{dj_hphb}%
\includegraphics[width=0.1\linewidth]{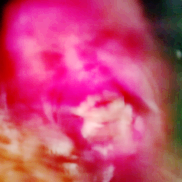}}%
\hspace{2mm}%
\subfigure[Bruce Willis]{%
\label{dj_hphb}%
\includegraphics[width=0.1\linewidth]{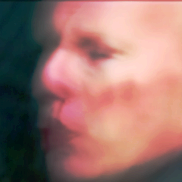}%
\label{dj_hphb}%
\includegraphics[width=0.1\linewidth]{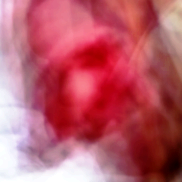}}%
\hspace{2mm}%
\subfigure[Chuck Norris]{%
\label{dj_hphb}%
\includegraphics[width=0.1\linewidth]{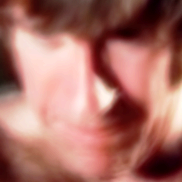}%
\label{dj_hphb}%
\includegraphics[width=0.1\linewidth]{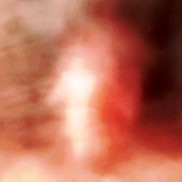}}%
\caption{Pair of average faces taken from sample shots (taken from low blurred and minimally occluded shots) of example subjects in Wildest Faces. Every first image represents a shot average with minimal pose variation and the second image is an average shot with severe pose variation. Comparison between these images indicates a large pose diversity in our dataset. Images are histogram equalized for convenience.}
\vspace{-4mm}
\label{pose_info}
\end{figure}

\noindent\textbf{Pose.} For selected individuals, we present four average faces (each taken from a shot). We make sure that there is no occlusion or high blur in these shots, so only pose variation is the concern.  It can be clearly seen from Figure~\ref{pose_info} that high pose variance leads to unidentifiable average faces supporting the complexity of Wildest Faces dateset.

\vspace{-4mm}

\section{Attentive Temporal Pooling} For the purpose of video face recognition, we propose a simple yet effective technique which we refer to as \textit{attentive temporal pooling}, inspired from \cite{yang2017neural}. The intuition behind this model is to exploit the hidden pose information in a trainable fashion to extract useful information in the noisy sequences of video frames. The proposed approach consists of three main components; i) an attention layer, ii) a pooling layer, and iii) a fully connected layer. Attention module learns to promote the informative parts of given image sequences. Through the pooling layer, the overall sequence information is aggregated and fed into a fully connected layer. This simple framework operates over CNN features.

More formally, the input $X$ is a $D \times F$ matrix of $D$-dimensional CNN feature vectors coming from $F$ frames. An attention weight matrix $A$ of size $K \times D$ is initialized using Xavier Normal Form method\cite{glorot2010understanding}. $K$ is a hyperparameter that needs to be tuned (we set $K=8$ in our experiments). The attention weights $S$ is calculated by
\begin{equation}
 S = A \times X
\end{equation}
\noindent which results in a $K \times F$ sized matrix. This matrix is then fed to a softmax function that operates over the temporal dimension. The $k$-th row of the resulting $K \times F$ matrix can be considered as a weight distribution over the frames, for the pose captured by the $k$-th row of the matrix $A$. We use the estimated attention weights to temporally pool the per-frame feature vectors. More specifically, we extract the video  feature vector by computing a weighted sum as follows:
\begin{equation}
O = X \times S^T
\label{eq:Odef}
\end{equation}
where the resulting matrix $O$ is of size $D \times K$. The output is then aggregated with max-pooling and fed into the fully connected layer which is used for classification with a cross-entropy loss. The model is implemented in PyTorch \cite{paszke2017pytorch}. The network parameters are optimized using SGD with a learning rate of 0.0001 and a momentum of 0.9. The batch size is set to 1. We note that this approach can be considered as a generalization of the aggregation scheme proposed in \cite{yang2017neural}, which is equivalent to Eq.~\ref{eq:Odef} for $K=1$.

\vspace{-4mm}
\section{Experimental Results} \label{results}

\subsection{Face Detection}

We first evaluate the performance of face detection over Wildest Faces dataset. For this purpose, we pick three most recent techniques; Single-Shot Scale Invariant Face Detector \cite{zhang2017s}, Tiny Faces \cite{hu2017finding} and Single Stage Headless Detector \cite{najibi2017ssh}. \footnote{We use the codes released by the papers' authors.}
We also evaluate a light-weight, SSD \cite{liu2016ssd}-based face detector available in OpenCV \footnote{\url{https://github.com/opencv/opencv/tree/master/samples/dnn/face_detector}}. We use all these techniques in an "as-is" configuration; we apply available pre-trained models (trained on WIDER Face \cite{yang2016wider}) on all our data (train, test and validation splits combined). Since our main focus in this work is on video face recognition, we do not perform any training on \textit{Wildest Faces}, hence we compute the performance of the detectors over the entire dataset of 67889 images. \vspace{2mm}

\begin{figure}
\centering
\subfigure[Severe blur.]{%
\label{severe_blur}%
\includegraphics[width=0.24\linewidth]{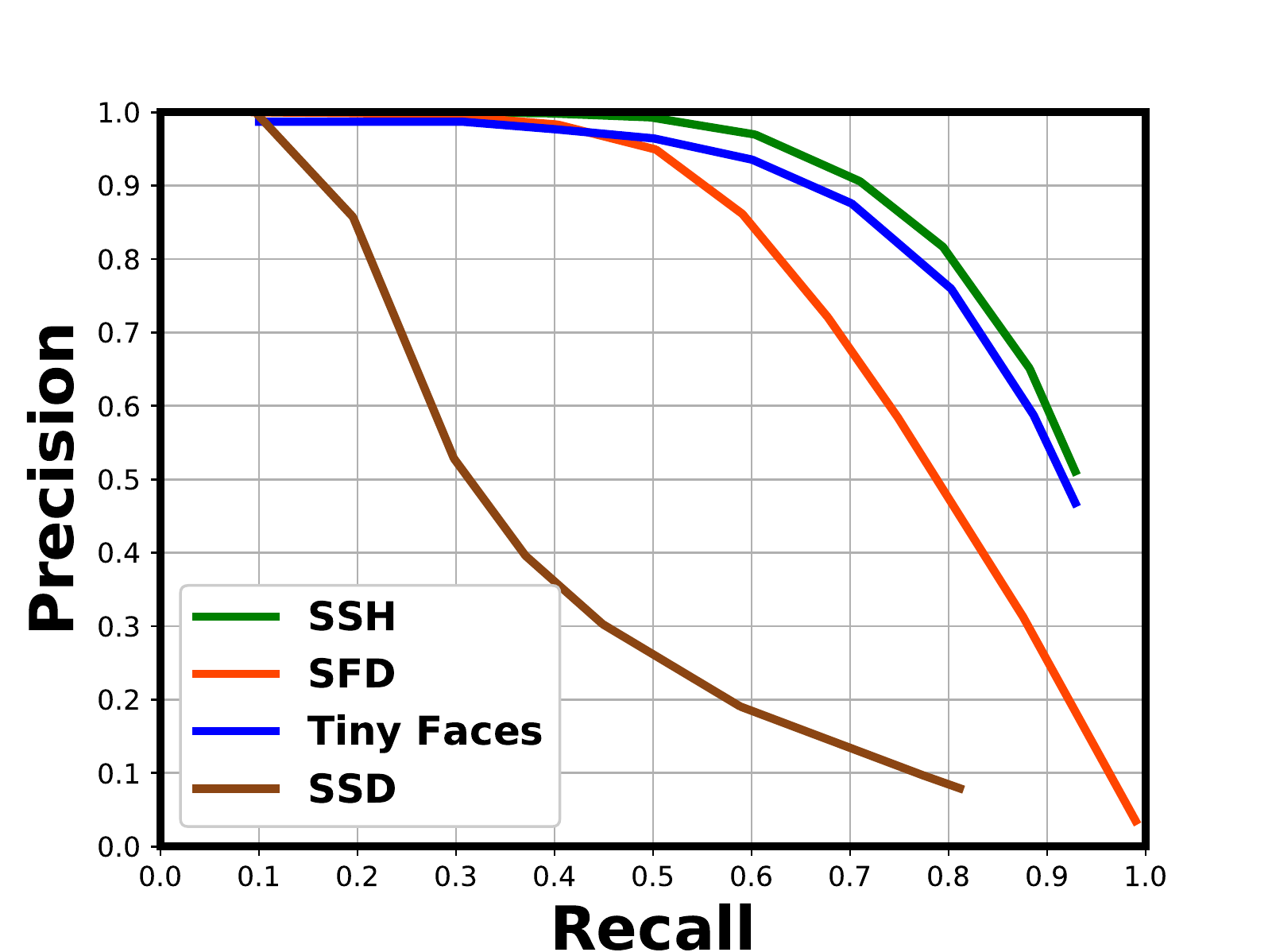}}%
\subfigure[High blur.]{%
\label{high_blur}%
\includegraphics[width=0.24\linewidth]{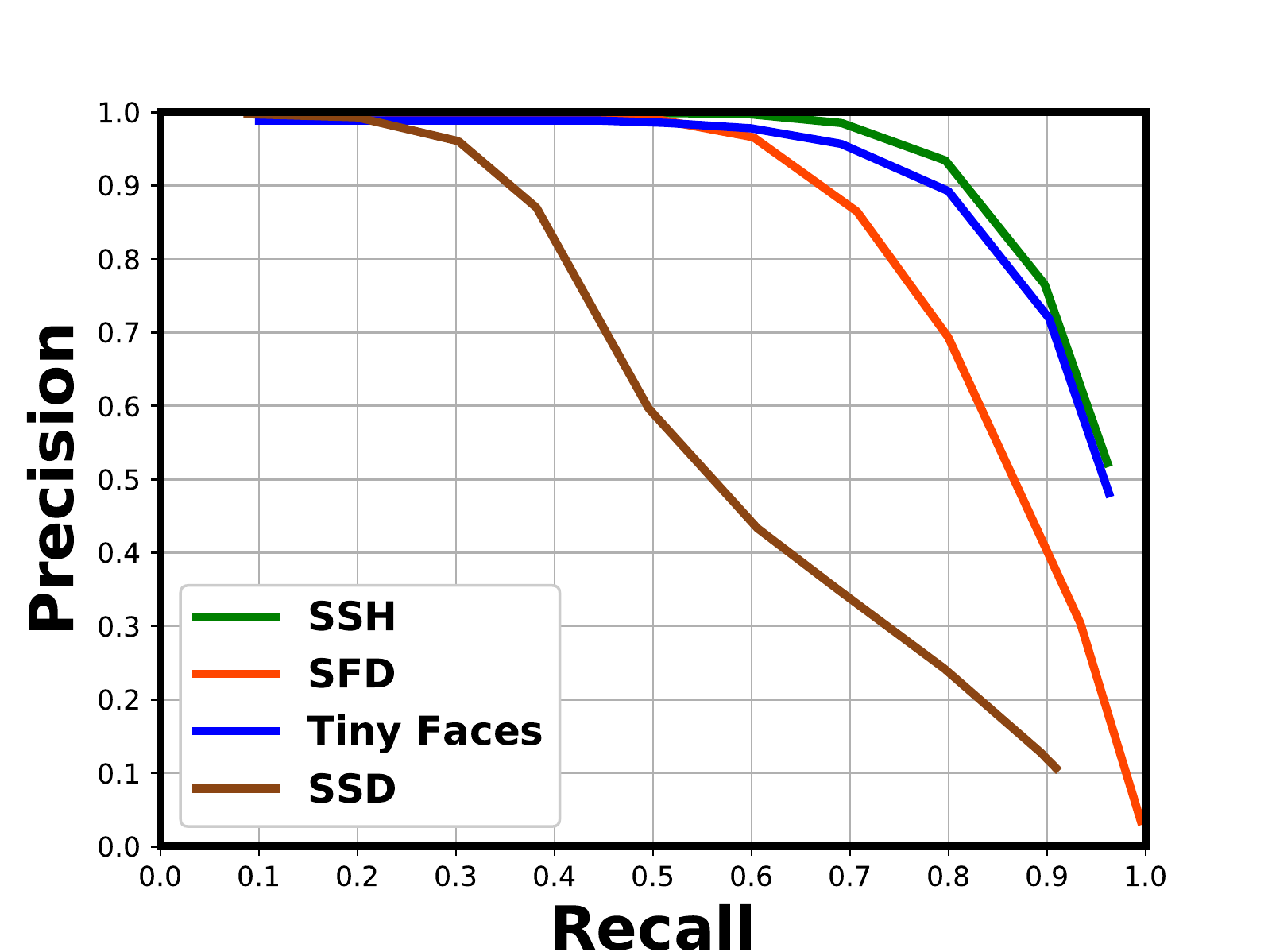}}%
\subfigure[Medium blur.]{%
\label{medium_blur}%
\includegraphics[width=0.24\linewidth]{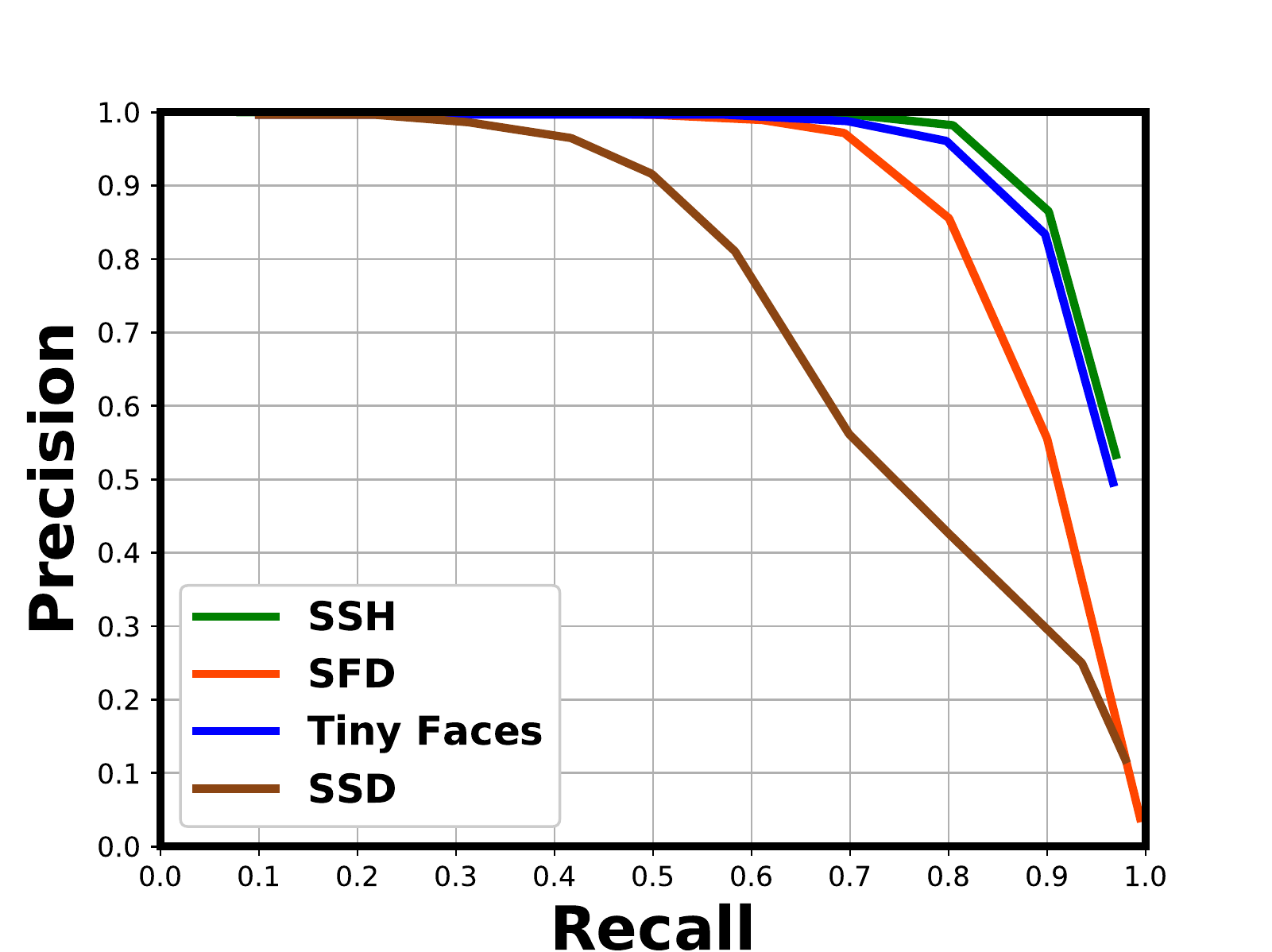}}%
\subfigure[Low blur.]{%
\label{low_blur}%
\includegraphics[width=0.24\linewidth]{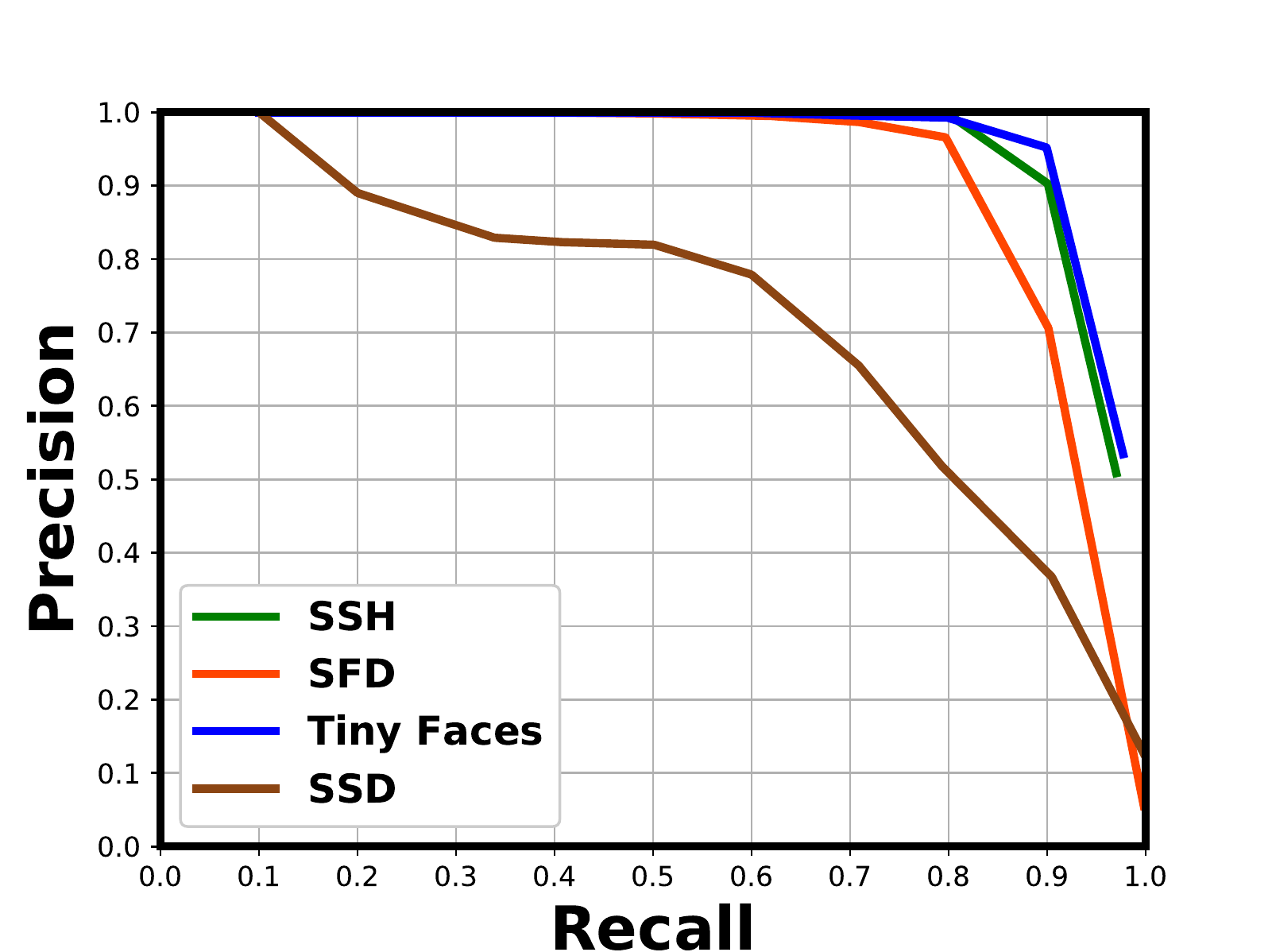}} \\
\vspace{-4mm} %
\subfigure[Large height.]{%
\label{large_height}%
\includegraphics[width=0.24\linewidth]{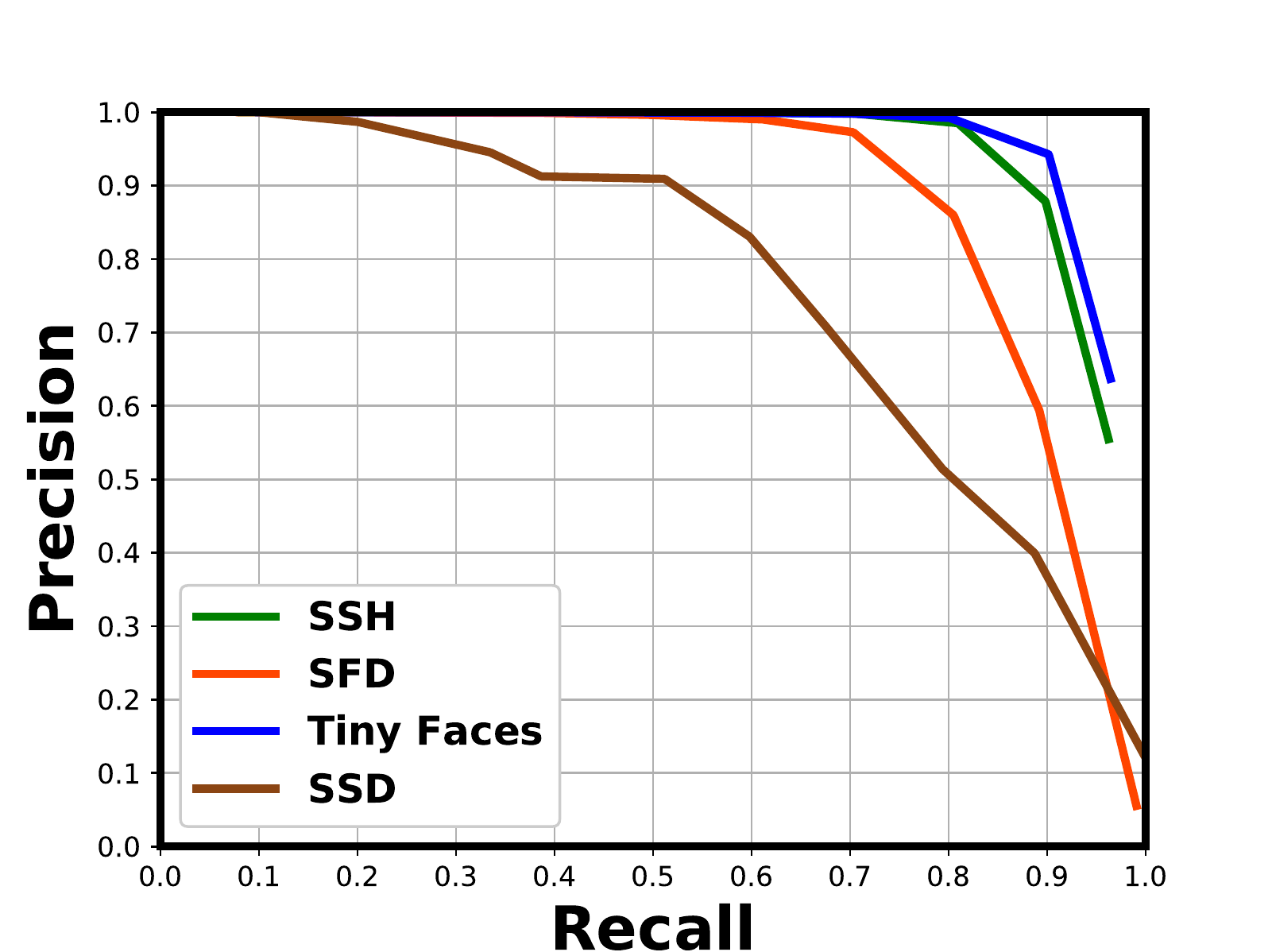}}%
\subfigure[Medium height.]{%
\label{medum_height}%
\includegraphics[width=0.24\linewidth]{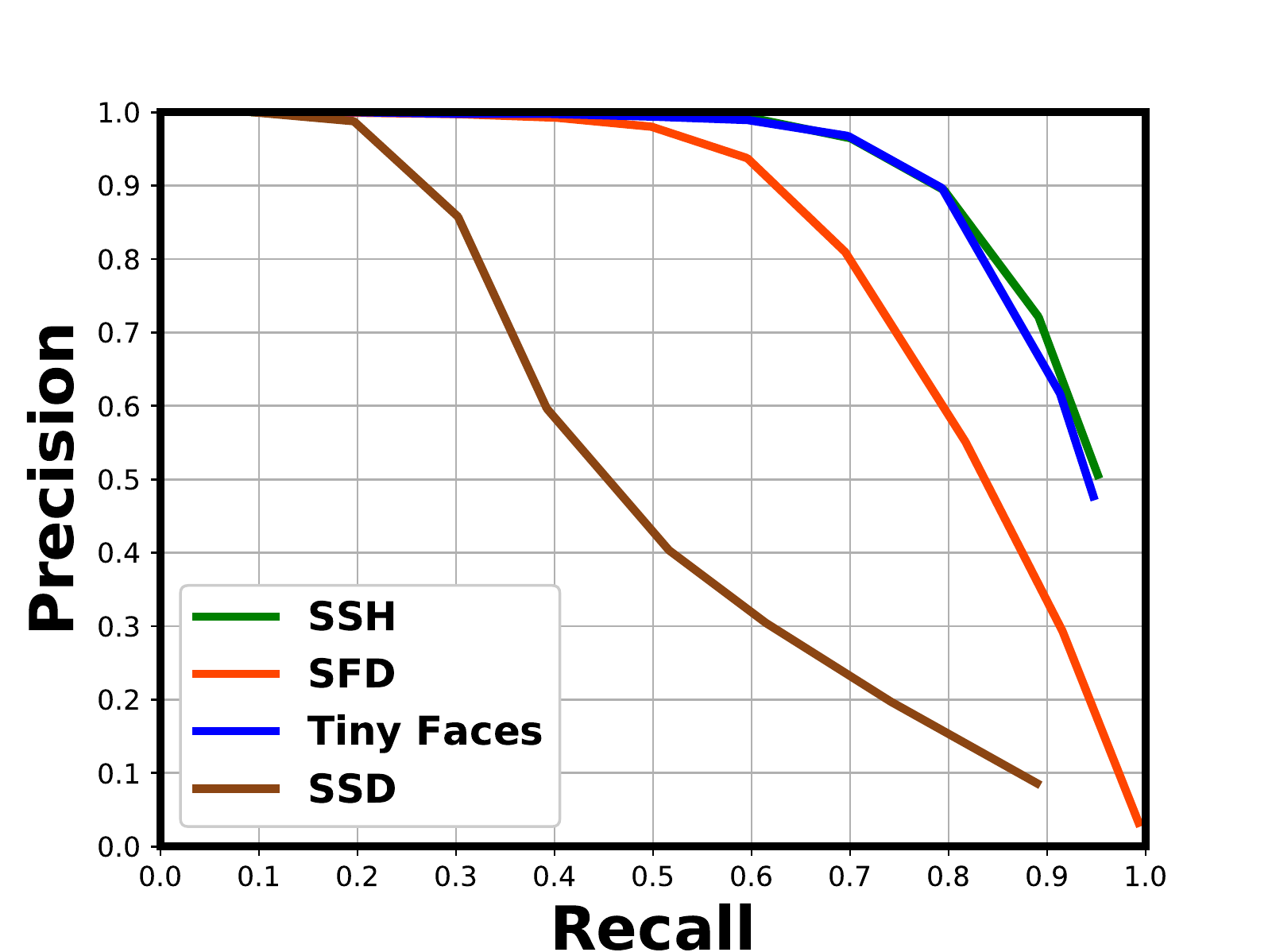}}%
\subfigure[Small height.]{%
\label{small_height}%
\includegraphics[width=0.24\linewidth]{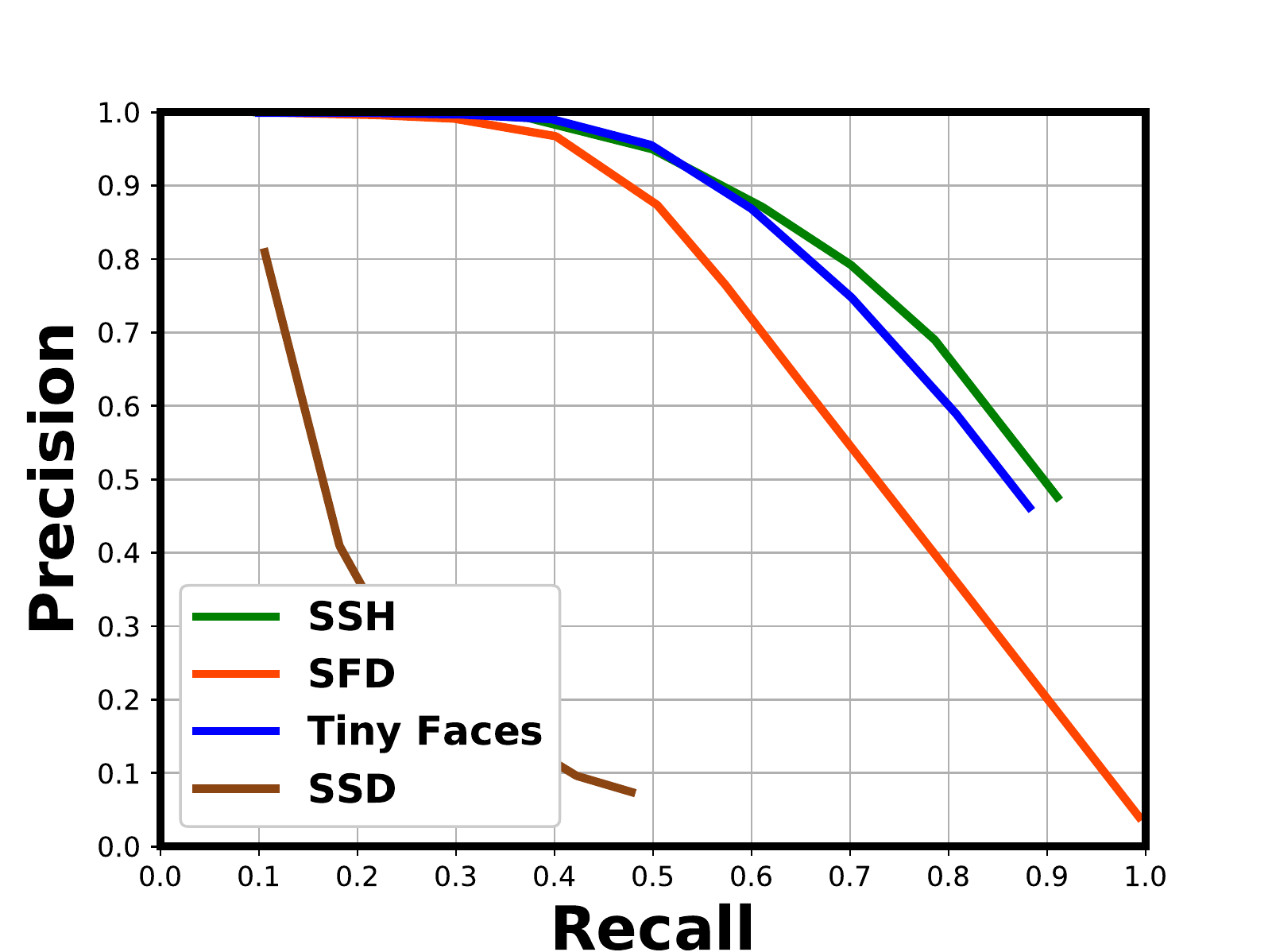}}%
\subfigure [Overall.]{%
\label{overall_pr}%
\includegraphics[width=0.24\linewidth]{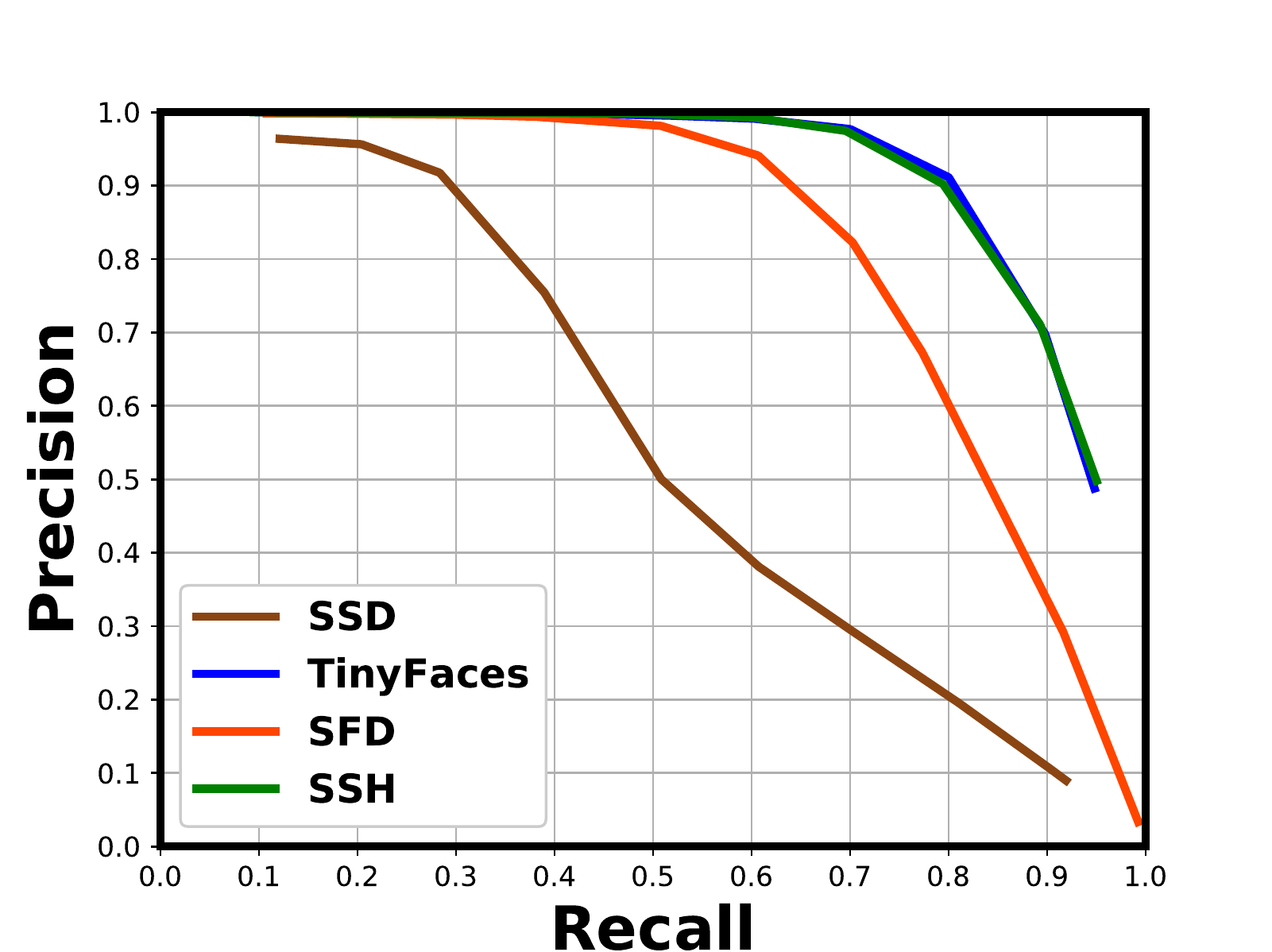}}%
\caption{Smaller scales and high blur levels severely degrade results of all face detectors.}
\label{detection_results}
\vspace{-6mm}
\end{figure}

\noindent\textbf{Overall.} Detection results are shown in Table \ref{det_ap} and Figure \ref{overall_pr}. It can be said that our dataset offers a new challenge for all the detectors. Performance-wise, we see Tiny Faces \cite{hu2017finding} and SSH \cite{najibi2017ssh} performing on par with each other. SFD \cite{zhang2017s} is the third best, whereas the light-weight SSD \cite{liu2016ssd} performs the worst.

\vspace{2mm}

\noindent\textbf{Blur.} Our blur analysis results are shown in Figure \ref{severe_blur} to \ref{low_blur}. We observe that blur severely degrades each detector; higher the blur worse the detection performance. SSH \cite{najibi2017ssh} seems to be the most robust detector to blur, whereas for low blur cases Tiny Faces \cite{hu2017finding} performs better with a slight margin. 

\vspace{2mm}

\noindent\textbf{Scale.} We test the performance of the detectors in different scales. Results are shown in Figure \ref{large_height} to \ref{small_height}. The same trend in overall performance is visible here as well; Tiny Faces\cite{hu2017finding} takes the lead over images with large size, with SSH \cite{najibi2017ssh} closely trailing behind, whereas the others fall visibly behind. As faces become smaller, SSH \cite{najibi2017ssh} catches up and takes the lead from Tiny Faces\cite{hu2017finding}. All the detectors have degraded performance when faces become smaller. We perform the same assessment for width and obtain a reminiscent trend.

These findings indicate that there is still considerable room for improvement for face detection in challenging cases like extreme blur or small size.

\vspace{-4mm}

\begin{table}[]
\normalsize
\centering

\resizebox{\textwidth}{!}{%
\begin{tabular}{l|l|l|l|l|l|l|l|l}
\hline
\textbf{Method}          & \textbf{Large} & \textbf{Medium } & \textbf{Small} & \textbf{Severe Blur} & \textbf{High Blur} & \textbf{Medium Blur} & \textbf{Low Blur} & \textbf{Overall} \\ \hline \hline
\textbf{SSD \cite{liu2016ssd}-based detector}                & 73.2\%            & 47.1\%            & 19.9\%            & 36\%              & 56.7\%           &     68\%        &       70.2\%        & 51.6\%  \\ \hline
\textbf{SFD \cite{zhang2017s}}                 & 84.6\%            & 75.9\%             & 69.5\%            & 74.3\%             & 78.4\%           & 84\%              & 87\%             & 77.3\%  \\ \hline

\textbf{Tiny Faces \cite{hu2017finding}}            &   \textbf{95.6\%}   & 89.3\%             & 80.7\%            & 85.2\%             & 89.6\%           & 92.5\%             & \textbf{94.6\%}   & 90.5\% \\ \hline
\textbf{SSH \cite{najibi2017ssh}}
   & 94.1\%            & \textbf{90.7\%}    & \textbf{82.4\%}   & \textbf{88.4\%}    & \textbf{92\%}   & \textbf{93.7\%}    & 94\%           & \textbf{90.7\%}  \\ \hline
\end{tabular}%
}
\vspace{-4mm}
\caption{Detection AP values. \textit{Small}, \textit{Medium} and \textit{Large} refer to height scale categories.}
\label{det_ap}
\vspace{-6mm}
\end{table}

\subsection{Face Recognition}

\subsubsection{Image-based Face Recognition}
For image-based face recognition, we use the train, validation and test splits of Wildest Faces dataset that consist of 39459, 12088, and  12695 face images respectively. We use two prominent face recognition approaches;  VGG Face  \cite{parkhi2015deep} and Center Loss\cite{wen2016discriminative} (trained on LFW \cite{huang2007labeled}). We first train these models from scratch over the \textit{Wildest Faces}, but we observe that they achieve significantly better results with pretrained models (trained on fairly larger datasets). We resize face regions to 96x96 and perform the relevant preprocessing steps in line with each technique's implementation using Caffe \cite{jia2014caffe}. We make minimal changes to original hyperparameters during training to improve convergence.

The image-based recognition results are shown in Table \ref{rec_results1}. Besides the comparison of face recognition techniques, we also test the effect of using alignment. For this purpose, we utilize  MTCNN alignment technique \cite{zhang2016joint}. We bypass the detector of MTCNN and use the ground-truth locations of faces during training. We add fully connected layers to the end of both networks of \cite{parkhi2015deep} and \cite{wen2016discriminative} to cast them as classifiers, since original models were for identification. The experimental results show that when no alignment is used, CenterLoss\cite{wen2016discriminative} method yields superior results. On the contrary, VGGFace\cite{parkhi2015deep} method benefits significantly from alignment and yields on par performance in the presence of alignment.

\vspace{-4mm}
\subsubsection{Video Face Recognition}
Our dataset consists of video clips of celebrities, so it is well-suited as a benchmark for video face recognition. The train, validation and test splits consist of 1347, 387 and 452 shots, respectively. The simplest baseline is majority voting using the techniques presented for standard face recognition. Results are shown in Table \ref{rec_results2}. We measure the recognition performance both at frame-level and at shot-level. Frame-level performance is evaluated as the accuracy over 12695 images and and shot-level over 452 shots, respectively.

For video face recognition, we also train several LSTM \cite{hochreiter1997long} architectures. Using the finetuned VGG features that are aligned with MTCNN, we implement single-layer LSTM, 2-layer LSTM (LSTM2), bi-directional LSTM (BiLSTM) and compare their performances with the attentive temporal pooling method described above. RMSprop optimizer with a learning rate of 0.0001 is used in all configurations of LSTMs for a fair comparison. Hidden sizes are fixed to 4096. Results are shown in Table \ref{rec_results2}. 

As expected, majority voting of standard image based techniques fails to yield competitive results at the shot-level, whereas frame-level accuracy of VGGFace\cite{parkhi2015deep} is competitive with video-based recognition techniques. At the shot-level, the best performing LSTM model is one-layer LSTM, whereas two-level LSTMs perform better at the shot-level. Overall, we observe that the proposed attentive temporal pooling model performs the best on average. Note that the accuracies are all around 50\% mark, indicating that the violent face recognition research can benefit from more tailored models. 

\begin{table}[]
\centering
\resizebox{0.45\textwidth}{!}{%
\begin{tabular}{l|l|l}
\hline
\textbf{Method}                      & \textbf{Alignment} & \textbf{Accuracy} \\ \hline \hline

\textbf{VGGFace \cite{parkhi2015deep}}                 & none  & 37.8\%   \\ 

\textbf{Center Loss  \cite{wen2016discriminative}}    & none         & 39.8\%  \\ 

\hline
\textbf{VGGFace \cite{parkhi2015deep}} &   \cite{zhang2016joint}          & 39.7\%    \\ 

\textbf{Center Loss \cite{wen2016discriminative}} &   \cite{zhang2016joint} & \textbf{39.9\%}          \\ \hline

\end{tabular}%
}
\caption{Image-based face recognition results. }
\label{rec_results1}
\vspace{-5mm}
\end{table}

\begin{table}[h]
\vspace{-2mm}
\centering
\resizebox{0.5\textwidth}{!}{%
\begin{tabular}{l|l|l}
\hline
\textbf{Method} & \textbf{Frame-Level}& \textbf{Shot-Level} \\
\hline\hline
\textbf{VGGFace\cite{parkhi2015deep}} & 51.98\%   &              49.5\%  \\
\textbf{CenterLoss\cite{wen2016discriminative}} &   49.6\%  &      46.6\%  \\

\textbf{LSTM} & 52.1\% &  51.9\%\\

\textbf{LSTM2} & \textbf{52.3\%} & 49.3\% \\
\textbf{BiLSTM} & 49.6\% & 50.6\%\\
\textbf{AttTempPool}  & 52.2\% & \textbf{52.6}\%\\
\hline
\end{tabular}
}
\caption{Accuracy values for video face recognition. In \textit{shot-level} evaluation, the accuracy is calculated over shots, whereas in \textit{frame-level}, accuracy is calculated over frames by assigning all the frames in the shot to the label of the sequence.}
\label{rec_results2}
\vspace{-5mm}
\end{table}

\vspace{-5mm}

\section{Conclusion} \label{conclude}
Inspired by the lack of a publicly available face detection and recognition dataset that concentrates primarily on violent scenes, we introduce \textit{Wildest Faces} dataset that compasses a large spectrum of adverse effects, such as severe blur, low resolution and a significant diversity in pose and occlusion. 
The dataset includes annotations for face detection as well as recognition with various tags, such as blur severity, scale and occlusion. To the best of our knowledge, this is the first  face dataset that focuses on violent scenes which inherently have extreme facial expressions along with challenging aspects.

We also provide benchmarks using prominent detection and recognition techniques and introduce an attention-based temporal pooling technique to aggregate video frames in a simple and effective way. We observe that approaches fall short to tackle the challenges of Wildest Faces. We hope Wildest Faces will boost face recognition and detection research towards edge cases. We will provide continuous improvements and additions to Wildest Faces dataset in the future.\footnote{The dataset with annotations will be made available upon publication}

\small
\bibliography{egbib}

\begin{thebibliography}{70}
\providecommand{\natexlab}[1]{#1}
\providecommand{\url}[1]{\texttt{#1}}
\expandafter\ifx\csname urlstyle\endcsname\relax
  \providecommand{\doi}[1]{doi: #1}\else
  \providecommand{\doi}{doi: \begingroup \urlstyle{rm}\Url}\fi

\bibitem[Ahonen et~al.(2006)Ahonen, Hadid, and Pietikainen]{ahonen2006face}
Timo Ahonen, Abdenour Hadid, and Matti Pietikainen.
\newblock Face description with local binary patterns: Application to face
  recognition.
\newblock \emph{IEEE transactions on pattern analysis and machine
  intelligence}, 28\penalty0 (12):\penalty0 2037--2041, 2006.

\bibitem[Beveridge et~al.(2013)Beveridge, Phillips, Bolme, Draper, Givens, Lui,
  Teli, Zhang, Scruggs, Bowyer, et~al.]{beveridge2013challenge}
J~Ross Beveridge, P~Jonathon Phillips, David~S Bolme, Bruce~A Draper, Geof~H
  Givens, Yui~Man Lui, Mohammad~Nayeem Teli, Hao Zhang, W~Todd Scruggs, Kevin~W
  Bowyer, et~al.
\newblock The challenge of face recognition from digital point-and-shoot
  cameras.
\newblock In \emph{Biometrics: Theory, Applications and Systems (BTAS), 2013
  IEEE Sixth International Conference on}, pages 1--8. IEEE, 2013.

\bibitem[Cao et~al.(2017)Cao, Shen, Xie, Parkhi, and
  Zisserman]{cao2017vggface2}
Qiong Cao, Li~Shen, Weidi Xie, Omkar~M Parkhi, and Andrew Zisserman.
\newblock Vggface2: A dataset for recognising faces across pose and age.
\newblock \emph{arXiv preprint arXiv:1710.08092}, 2017.

\bibitem[Cheng et~al.(2018)Cheng, Zhou, and Han]{cheng2018duplex}
Gong Cheng, Peicheng Zhou, and Junwei Han.
\newblock Duplex metric learning for image set classification.
\newblock \emph{IEEE Transactions on Image Processing}, 27\penalty0
  (1):\penalty0 281--292, 2018.

\bibitem[Chowdhury et~al.(2016)Chowdhury, Lin, Maji, and
  Learned-Miller]{chowdhury2016one}
Aruni~Roy Chowdhury, Tsung-Yu Lin, Subhransu Maji, and Erik Learned-Miller.
\newblock One-to-many face recognition with bilinear cnns.
\newblock In \emph{Applications of Computer Vision (WACV), 2016 IEEE Winter
  Conference on}, pages 1--9. IEEE, 2016.

\bibitem[Ding and Tao(2016)]{ding2016comprehensive}
Changxing Ding and Dacheng Tao.
\newblock A comprehensive survey on pose-invariant face recognition.
\newblock \emph{ACM Transactions on intelligent systems and technology (TIST)},
  7\penalty0 (3):\penalty0 37, 2016.

\bibitem[Edwards et~al.(1998)Edwards, Cootes, and Taylor]{edwards1998face}
Gareth~J Edwards, Timothy~F Cootes, and Christopher~J Taylor.
\newblock Face recognition using active appearance models.
\newblock In \emph{European conference on computer vision}, pages 581--595.
  Springer, 1998.

\bibitem[Everingham et~al.(2010)Everingham, Van~Gool, Williams, Winn, and
  Zisserman]{everingham2010pascal}
Mark Everingham, Luc Van~Gool, Christopher~KI Williams, John Winn, and Andrew
  Zisserman.
\newblock The pascal visual object classes (voc) challenge.
\newblock \emph{International journal of computer vision}, 88\penalty0
  (2):\penalty0 303--338, 2010.

\bibitem[Farfade et~al.(2015)Farfade, Saberian, and Li]{farfade2015multi}
Sachin~Sudhakar Farfade, Mohammad~J Saberian, and Li-Jia Li.
\newblock Multi-view face detection using deep convolutional neural networks.
\newblock In \emph{Proceedings of the 5th ACM on International Conference on
  Multimedia Retrieval}, pages 643--650. ACM, 2015.

\bibitem[Glorot and Bengio(2010)]{glorot2010understanding}
Xavier Glorot and Yoshua Bengio.
\newblock Understanding the difficulty of training deep feedforward neural
  networks.
\newblock In \emph{Proceedings of the thirteenth international conference on
  artificial intelligence and statistics}, pages 249--256, 2010.

\bibitem[Goswami et~al.(2014)Goswami, Bhardwaj, Singh, and
  Vatsa]{goswami2014mdlface}
Gaurav Goswami, Romil Bhardwaj, Richa Singh, and Mayank Vatsa.
\newblock Mdlface: Memorability augmented deep learning for video face
  recognition.
\newblock In \emph{Biometrics (IJCB), 2014 IEEE International Joint Conference
  on}, pages 1--7. IEEE, 2014.

\bibitem[Goswami et~al.(2017)Goswami, Vatsa, and Singh]{goswami2017face}
Gaurav Goswami, Mayank Vatsa, and Richa Singh.
\newblock Face verification via learned representation on feature-rich video
  frames.
\newblock \emph{IEEE Transactions on Information Forensics and Security},
  12\penalty0 (7):\penalty0 1686--1698, 2017.

\bibitem[Guo et~al.(2016)Guo, Zhang, Hu, He, and Gao]{guo2016ms}
Yandong Guo, Lei Zhang, Yuxiao Hu, Xiaodong He, and Jianfeng Gao.
\newblock Ms-celeb-1m: Challenge of recognizing one million celebrities in the
  real world.
\newblock \emph{Electronic Imaging}, 2016\penalty0 (11):\penalty0 1--6, 2016.

\bibitem[Hochreiter and Schmidhuber(1997)]{hochreiter1997long}
Sepp Hochreiter and J{\"u}rgen Schmidhuber.
\newblock Long short-term memory.
\newblock \emph{Neural computation}, 9\penalty0 (8):\penalty0 1735--1780, 1997.

\bibitem[Hu and Ramanan(2017)]{hu2017finding}
Peiyun Hu and Deva Ramanan.
\newblock Finding tiny faces.
\newblock In \emph{2017 IEEE Conference on Computer Vision and Pattern
  Recognition (CVPR)}, pages 1522--1530. IEEE, 2017.

\bibitem[Huang et~al.(2007)Huang, Ramesh, Berg, and
  Learned-Miller]{huang2007labeled}
Gary~B Huang, Manu Ramesh, Tamara Berg, and Erik Learned-Miller.
\newblock Labeled faces in the wild: A database for studying face recognition
  in unconstrained environments.
\newblock Technical report, Technical Report 07-49, University of
  Massachusetts, Amherst, 2007.

\bibitem[Huang et~al.(2015{\natexlab{a}})Huang, Shan, Wang, Zhang, Lao,
  Kuerban, and Chen]{huang2015benchmark}
Zhiwu Huang, Shiguang Shan, Ruiping Wang, Haihong Zhang, Shihong Lao, Alifu
  Kuerban, and Xilin Chen.
\newblock A benchmark and comparative study of video-based face recognition on
  cox face database.
\newblock \emph{IEEE Transactions on Image Processing}, 24\penalty0
  (12):\penalty0 5967--5981, 2015{\natexlab{a}}.

\bibitem[Huang et~al.(2015{\natexlab{b}})Huang, Wang, Shan, Li, and
  Chen]{huang2015log}
Zhiwu Huang, Ruiping Wang, Shiguang Shan, Xianqiu Li, and Xilin Chen.
\newblock Log-euclidean metric learning on symmetric positive definite manifold
  with application to image set classification.
\newblock In \emph{International conference on machine learning}, pages
  720--729, 2015{\natexlab{b}}.

\bibitem[Huang et~al.(2017)Huang, Wang, Van~Gool, Chen, et~al.]{huang2017cross}
Zhiwu Huang, Ruiping Wang, Luc Van~Gool, Xilin Chen, et~al.
\newblock Cross euclidean-to-riemannian metric learning with application to
  face recognition from video.
\newblock \emph{IEEE Transactions on Pattern Analysis and Machine
  Intelligence}, 2017.

\bibitem[Jain and Learned-Miller(2010)]{jain2010fddb}
Vidit Jain and Erik Learned-Miller.
\newblock Fddb: A benchmark for face detection in unconstrained settings.
\newblock \emph{University of Massachusetts, Amherst, Tech. Rep.
  UM-CS-2010-009}, 2\penalty0 (7):\penalty0 8, 2010.

\bibitem[Jia et~al.(2014)Jia, Shelhamer, Donahue, Karayev, Long, Girshick,
  Guadarrama, and Darrell]{jia2014caffe}
Yangqing Jia, Evan Shelhamer, Jeff Donahue, Sergey Karayev, Jonathan Long, Ross
  Girshick, Sergio Guadarrama, and Trevor Darrell.
\newblock Caffe: Convolutional architecture for fast feature embedding.
\newblock In \emph{Proceedings of the 22nd ACM international conference on
  Multimedia}, pages 675--678. ACM, 2014.

\bibitem[Kemelmacher-Shlizerman et~al.(2016)Kemelmacher-Shlizerman, Seitz,
  Miller, and Brossard]{kemelmacher2016megaface}
Ira Kemelmacher-Shlizerman, Steven~M Seitz, Daniel Miller, and Evan Brossard.
\newblock The megaface benchmark: 1 million faces for recognition at scale.
\newblock In \emph{Proceedings of the IEEE Conference on Computer Vision and
  Pattern Recognition}, pages 4873--4882, 2016.

\bibitem[Klare et~al.(2015)Klare, Klein, Taborsky, Blanton, Cheney, Allen,
  Grother, Mah, and Jain]{klare2015pushing}
Brendan~F Klare, Ben Klein, Emma Taborsky, Austin Blanton, Jordan Cheney,
  Kristen Allen, Patrick Grother, Alan Mah, and Anil~K Jain.
\newblock Pushing the frontiers of unconstrained face detection and
  recognition: Iarpa janus benchmark a.
\newblock In \emph{Proceedings of the IEEE Conference on Computer Vision and
  Pattern Recognition}, pages 1931--1939, 2015.

\bibitem[Kumar et~al.(2009)Kumar, Berg, Belhumeur, and
  Nayar]{kumar2009attribute}
Neeraj Kumar, Alexander~C Berg, Peter~N Belhumeur, and Shree~K Nayar.
\newblock Attribute and simile classifiers for face verification.
\newblock In \emph{Computer Vision, 2009 IEEE 12th International Conference
  on}, pages 365--372. IEEE, 2009.

\bibitem[Li et~al.(2013)Li, Hua, Lin, Brandt, and Yang]{li2013probabilistic}
Haoxiang Li, Gang Hua, Zhe Lin, Jonathan Brandt, and Jianchao Yang.
\newblock Probabilistic elastic matching for pose variant face verification.
\newblock In \emph{Computer Vision and Pattern Recognition (CVPR), 2013 IEEE
  Conference on}, pages 3499--3506. IEEE, 2013.

\bibitem[Li et~al.(2014)Li, Hua, Shen, Lin, and Brandt]{li2014eigen}
Haoxiang Li, Gang Hua, Xiaohui Shen, Zhe Lin, and Jonathan Brandt.
\newblock Eigen-pep for video face recognition.
\newblock In \emph{Asian Conference on Computer Vision}, pages 17--33.
  Springer, 2014.

\bibitem[Li et~al.(2015)Li, Lin, Shen, Brandt, and Hua]{li2015convolutional}
Haoxiang Li, Zhe Lin, Xiaohui Shen, Jonathan Brandt, and Gang Hua.
\newblock A convolutional neural network cascade for face detection.
\newblock In \emph{Proceedings of the IEEE Conference on Computer Vision and
  Pattern Recognition}, pages 5325--5334, 2015.

\bibitem[Li and Zhang(2013)]{li2013learning}
Jianguo Li and Yimin Zhang.
\newblock Learning surf cascade for fast and accurate object detection.
\newblock In \emph{Computer Vision and Pattern Recognition (CVPR), 2013 IEEE
  Conference on}, pages 3468--3475. IEEE, 2013.

\bibitem[Liu et~al.(2016)Liu, Anguelov, Erhan, Szegedy, Reed, Fu, and
  Berg]{liu2016ssd}
Wei Liu, Dragomir Anguelov, Dumitru Erhan, Christian Szegedy, Scott Reed,
  Cheng-Yang Fu, and Alexander~C Berg.
\newblock Ssd: Single shot multibox detector.
\newblock In \emph{European conference on computer vision}, pages 21--37.
  Springer, 2016.

\bibitem[Martin~Koestinger and Bischof(2011)]{koestinger11a}
Peter M.~Roth Martin~Koestinger, Paul~Wohlhart and Horst Bischof.
\newblock {Annotated Facial Landmarks in the Wild: A Large-scale, Real-world
  Database for Facial Landmark Localization}.
\newblock In \emph{{Proc. First IEEE International Workshop on Benchmarking
  Facial Image Analysis Technologies}}, 2011.

\bibitem[Mathias et~al.(2014)Mathias, Benenson, Pedersoli, and
  Van~Gool]{mathias2014face}
Markus Mathias, Rodrigo Benenson, Marco Pedersoli, and Luc Van~Gool.
\newblock Face detection without bells and whistles.
\newblock In \emph{European Conference on Computer Vision}, pages 720--735.
  Springer, 2014.

\bibitem[Najibi et~al.(2017)Najibi, Samangouei, Chellappa, and
  Davis]{najibi2017ssh}
Mahyar Najibi, Pouya Samangouei, Rama Chellappa, and Larry Davis.
\newblock Ssh: Single stage headless face detector.
\newblock In \emph{Proceedings of the IEEE Conference on Computer Vision and
  Pattern Recognition}, pages 4875--4884, 2017.

\bibitem[Naphade et~al.(2006)Naphade, Smith, Tesic, Chang, Hsu, Kennedy,
  Hauptmann, and Curtis]{naphade2006large}
Milind Naphade, John~R Smith, Jelena Tesic, Shih-Fu Chang, Winston Hsu, Lyndon
  Kennedy, Alexander Hauptmann, and Jon Curtis.
\newblock Large-scale concept ontology for multimedia.
\newblock \emph{IEEE multimedia}, 13\penalty0 (3):\penalty0 86--91, 2006.

\bibitem[Nech and Kemelmacher-Shlizerman(2017)]{nech2017level}
Aaron Nech and Ira Kemelmacher-Shlizerman.
\newblock Level playing field for million scale face recognition.
\newblock In \emph{2017 IEEE Conference on Computer Vision and Pattern
  Recognition (CVPR)}, pages 3406--3415. IEEE, 2017.

\bibitem[Ng and Winkler(2014)]{ng2014data}
Hong-Wei Ng and Stefan Winkler.
\newblock A data-driven approach to cleaning large face datasets.
\newblock In \emph{Image Processing (ICIP), 2014 IEEE International Conference
  on}, pages 343--347. IEEE, 2014.

\bibitem[Parkhi et~al.(2014)Parkhi, Simonyan, Vedaldi, and
  Zisserman]{parkhi2014compact}
Omkar~M Parkhi, Karen Simonyan, Andrea Vedaldi, and Andrew Zisserman.
\newblock A compact and discriminative face track descriptor.
\newblock In \emph{Proceedings of the IEEE Conference on Computer Vision and
  Pattern Recognition}, pages 1693--1700, 2014.

\bibitem[Parkhi et~al.(2015)Parkhi, Vedaldi, Zisserman, et~al.]{parkhi2015deep}
Omkar~M Parkhi, Andrea Vedaldi, Andrew Zisserman, et~al.
\newblock Deep face recognition.
\newblock In \emph{BMVC}, volume~1, page~6, 2015.

\bibitem[Paszke et~al.(2017)Paszke, Gross, Chintala, and
  Chanan]{paszke2017pytorch}
Adam Paszke, Sam Gross, Soumith Chintala, and Gregory Chanan.
\newblock Pytorch.
\newblock 2017.

\bibitem[Pech-Pacheco et~al.(2000)Pech-Pacheco, Crist{\'o}bal,
  Chamorro-Martinez, and Fern{\'a}ndez-Valdivia]{pech2000diatom}
Jos{\'e}~Luis Pech-Pacheco, Gabriel Crist{\'o}bal, Jes{\'u}s Chamorro-Martinez,
  and Joaqu{\'\i}n Fern{\'a}ndez-Valdivia.
\newblock Diatom autofocusing in brightfield microscopy: a comparative study.
\newblock In \emph{Pattern Recognition, 2000. Proceedings. 15th International
  Conference on}, volume~3, pages 314--317. IEEE, 2000.

\bibitem[Rao et~al.(2017{\natexlab{a}})Rao, Lin, Lu, and Zhou]{rao2017learning}
Yongming Rao, Ji~Lin, Jiwen Lu, and Jie Zhou.
\newblock Learning discriminative aggregation network for video-based face
  recognition.
\newblock In \emph{Proceedings of the IEEE Conference on Computer Vision and
  Pattern Recognition}, pages 3781--3790, 2017{\natexlab{a}}.

\bibitem[Rao et~al.(2017{\natexlab{b}})Rao, Lu, and Zhou]{rao2017attention}
Yongming Rao, Jiwen Lu, and Jie Zhou.
\newblock Attention-aware deep reinforcement learning for video face
  recognition.
\newblock In \emph{Proceedings of the IEEE Conference on Computer Vision and
  Pattern Recognition}, pages 3931--3940, 2017{\natexlab{b}}.

\bibitem[Samangouei et~al.(2018)Samangouei, Najibi, Davis, and
  Chellappa]{samangouei2018face}
Pouya Samangouei, Mahyar Najibi, Larry Davis, and Rama Chellappa.
\newblock Face-magnet: Magnifying feature maps to detect small faces.
\newblock \emph{arXiv preprint arXiv:1803.05258}, 2018.

\bibitem[Schroff et~al.(2015)Schroff, Kalenichenko, and
  Philbin]{schroff2015facenet}
Florian Schroff, Dmitry Kalenichenko, and James Philbin.
\newblock Facenet: A unified embedding for face recognition and clustering.
\newblock In \emph{Proceedings of the IEEE conference on computer vision and
  pattern recognition}, pages 815--823, 2015.

\bibitem[Sun et~al.(2013)Sun, Wang, and Tang]{sun2013hybrid}
Yi~Sun, Xiaogang Wang, and Xiaoou Tang.
\newblock Hybrid deep learning for face verification.
\newblock In \emph{Computer Vision (ICCV), 2013 IEEE International Conference
  on}, pages 1489--1496. IEEE, 2013.

\bibitem[Sun et~al.(2014)Sun, Chen, Wang, and Tang]{sun2014deep}
Yi~Sun, Yuheng Chen, Xiaogang Wang, and Xiaoou Tang.
\newblock Deep learning face representation by joint
  identification-verification.
\newblock In \emph{Advances in neural information processing systems}, pages
  1988--1996, 2014.

\bibitem[Sun et~al.(2015)Sun, Liang, Wang, and Tang]{sun2015deepid3}
Yi~Sun, Ding Liang, Xiaogang Wang, and Xiaoou Tang.
\newblock Deepid3: Face recognition with very deep neural networks.
\newblock \emph{arXiv preprint arXiv:1502.00873}, 2015.

\bibitem[Taigman et~al.(2014)Taigman, Yang, Ranzato, and
  Wolf]{taigman2014deepface}
Yaniv Taigman, Ming Yang, Marc'Aurelio Ranzato, and Lior Wolf.
\newblock Deepface: Closing the gap to human-level performance in face
  verification.
\newblock In \emph{Proceedings of the IEEE conference on computer vision and
  pattern recognition}, pages 1701--1708, 2014.

\bibitem[Tang et~al.(2018)Tang, Du, He, and Liu]{tang2018pyramidbox}
Xu~Tang, Daniel~K Du, Zeqiang He, and Jingtuo Liu.
\newblock Pyramidbox: A context-assisted single shot face detector.
\newblock \emph{arXiv preprint arXiv:1803.07737}, 2018.

\bibitem[Turk and Pentland(1991)]{turk1991face}
Matthew~A Turk and Alex~P Pentland.
\newblock Face recognition using eigenfaces.
\newblock In \emph{Computer Vision and Pattern Recognition, 1991. Proceedings
  CVPR'91., IEEE Computer Society Conference on}, pages 586--591. IEEE, 1991.

\bibitem[Viola and Jones(2001)]{viola2001rapid}
Paul Viola and Michael Jones.
\newblock Rapid object detection using a boosted cascade of simple features.
\newblock In \emph{Computer Vision and Pattern Recognition, 2001. CVPR 2001.
  Proceedings of the 2001 IEEE Computer Society Conference on}, volume~1, pages
  I--I. IEEE, 2001.

\bibitem[Viola and Jones(2004)]{viola2004robust}
Paul Viola and Michael~J Jones.
\newblock Robust real-time face detection.
\newblock \emph{International journal of computer vision}, 57\penalty0
  (2):\penalty0 137--154, 2004.

\bibitem[Wang et~al.(2017)Wang, Yuan, and Yu]{wang2017face}
Jianfeng Wang, Ye~Yuan, and Gang Yu.
\newblock Face attention network: An effective face detector for the occluded
  faces.
\newblock \emph{arXiv preprint arXiv:1711.07246}, 2017.

\bibitem[Wen et~al.(2016)Wen, Zhang, Li, and Qiao]{wen2016discriminative}
Yandong Wen, Kaipeng Zhang, Zhifeng Li, and Yu~Qiao.
\newblock A discriminative feature learning approach for deep face recognition.
\newblock In \emph{European Conference on Computer Vision}, pages 499--515.
  Springer, 2016.

\bibitem[Wiskott et~al.(1997)Wiskott, Kr{\"u}ger, Kuiger, and Von
  Der~Malsburg]{wiskott1997face}
Laurenz Wiskott, Norbert Kr{\"u}ger, N~Kuiger, and Christoph Von Der~Malsburg.
\newblock Face recognition by elastic bunch graph matching.
\newblock \emph{IEEE Transactions on pattern analysis and machine
  intelligence}, 19\penalty0 (7):\penalty0 775--779, 1997.

\bibitem[Wolf et~al.(2011)Wolf, Hassner, and Maoz]{wolf2011face}
Lior Wolf, Tal Hassner, and Itay Maoz.
\newblock Face recognition in unconstrained videos with matched background
  similarity.
\newblock In \emph{Computer Vision and Pattern Recognition (CVPR), 2011 IEEE
  Conference on}, pages 529--534. IEEE, 2011.

\bibitem[Wright et~al.(2009)Wright, Yang, Ganesh, Sastry, and
  Ma]{wright2009robust}
John Wright, Allen~Y Yang, Arvind Ganesh, S~Shankar Sastry, and Yi~Ma.
\newblock Robust face recognition via sparse representation.
\newblock \emph{IEEE transactions on pattern analysis and machine
  intelligence}, 31\penalty0 (2):\penalty0 210--227, 2009.

\bibitem[Xie et~al.(2010)Xie, Shan, Chen, and Chen]{xie2010fusing}
Shufu Xie, Shiguang Shan, Xilin Chen, and Jie Chen.
\newblock Fusing local patterns of gabor magnitude and phase for face
  recognition.
\newblock \emph{IEEE transactions on image processing}, 19\penalty0
  (5):\penalty0 1349--1361, 2010.

\bibitem[Yan et~al.(2014)Yan, Zhang, Lei, and Li]{yan2014face}
Junjie Yan, Xuzong Zhang, Zhen Lei, and Stan~Z Li.
\newblock Face detection by structural models.
\newblock \emph{Image and Vision Computing}, 32\penalty0 (10):\penalty0
  790--799, 2014.

\bibitem[Yang et~al.(2014)Yang, Yan, Lei, and Li]{yang2014aggregate}
Bin Yang, Junjie Yan, Zhen Lei, and Stan~Z Li.
\newblock Aggregate channel features for multi-view face detection.
\newblock In \emph{Biometrics (IJCB), 2014 IEEE International Joint Conference
  on}, pages 1--8. IEEE, 2014.

\bibitem[Yang et~al.(2015{\natexlab{a}})Yang, Yan, Lei, and Li]{yang2015fine}
Bin Yang, Junjie Yan, Zhen Lei, and Stan~Z Li.
\newblock Fine-grained evaluation on face detection in the wild.
\newblock In \emph{Automatic Face and Gesture Recognition (FG), 2015 11th IEEE
  International Conference and Workshops on}, volume~1, pages 1--7. IEEE,
  2015{\natexlab{a}}.

\bibitem[Yang et~al.(2017{\natexlab{a}})Yang, Ren, Chen, Wen, Li, and
  Hua]{yang2017neural}
Jiaolong Yang, Peiran Ren, Dong Chen, Fang Wen, Hongdong Li, and Gang Hua.
\newblock Neural aggregation network for video face recognition.
\newblock \emph{arXiv preprint}, 2017{\natexlab{a}}.

\bibitem[Yang et~al.(2015{\natexlab{b}})Yang, Luo, Loy, and
  Tang]{yang2015facial}
Shuo Yang, Ping Luo, Chen-Change Loy, and Xiaoou Tang.
\newblock From facial parts responses to face detection: A deep learning
  approach.
\newblock In \emph{Proceedings of the IEEE International Conference on Computer
  Vision}, pages 3676--3684, 2015{\natexlab{b}}.

\bibitem[Yang et~al.(2016)Yang, Luo, Loy, and Tang]{yang2016wider}
Shuo Yang, Ping Luo, Chen-Change Loy, and Xiaoou Tang.
\newblock Wider face: A face detection benchmark.
\newblock In \emph{Proceedings of the IEEE Conference on Computer Vision and
  Pattern Recognition}, pages 5525--5533, 2016.

\bibitem[Yang et~al.(2017{\natexlab{b}})Yang, Xiong, Loy, and
  Tang]{yang2017face}
Shuo Yang, Yuanjun Xiong, Chen~Change Loy, and Xiaoou Tang.
\newblock Face detection through scale-friendly deep convolutional networks.
\newblock \emph{arXiv preprint arXiv:1706.02863}, 2017{\natexlab{b}}.

\bibitem[Yi et~al.(2014)Yi, Lei, Liao, and Li]{yi2014learning}
Dong Yi, Zhen Lei, Shengcai Liao, and Stan~Z Li.
\newblock Learning face representation from scratch.
\newblock \emph{arXiv preprint arXiv:1411.7923}, 2014.

\bibitem[Zafeiriou et~al.(2015)Zafeiriou, Zhang, and
  Zhang]{zafeiriou2015survey}
Stefanos Zafeiriou, Cha Zhang, and Zhengyou Zhang.
\newblock A survey on face detection in the wild: past, present and future.
\newblock \emph{Computer Vision and Image Understanding}, 138:\penalty0 1--24,
  2015.

\bibitem[Zhang et~al.(2016)Zhang, Zhang, Li, and Qiao]{zhang2016joint}
Kaipeng Zhang, Zhanpeng Zhang, Zhifeng Li, and Yu~Qiao.
\newblock Joint face detection and alignment using multitask cascaded
  convolutional networks.
\newblock \emph{IEEE Signal Processing Letters}, 23\penalty0 (10):\penalty0
  1499--1503, 2016.

\bibitem[Zhang et~al.(2017)Zhang, Zhu, Lei, Shi, Wang, and Li]{zhang2017s}
Shifeng Zhang, Xiangyu Zhu, Zhen Lei, Hailin Shi, Xiaobo Wang, and Stan~Z Li.
\newblock Single shot scale-invariant face detector.
\newblock \emph{arXiv preprint arXiv:1708.05237}, 2017.

\bibitem[Zhu et~al.(2018)Zhu, Tao, Luu, and Savvides]{zhu2018seeing}
Chenchen Zhu, Ran Tao, Khoa Luu, and Marios Savvides.
\newblock Seeing small faces from robust anchor's perspective.
\newblock \emph{arXiv preprint arXiv:1802.09058}, 2018.

\bibitem[Zhu and Ramanan(2012)]{zhu2012face}
Xiangxin Zhu and Deva Ramanan.
\newblock Face detection, pose estimation, and landmark localization in the
  wild.
\newblock In \emph{Computer Vision and Pattern Recognition (CVPR), 2012 IEEE
  Conference on}, pages 2879--2886. IEEE, 2012.

\end{thebibliography}
\end{document}